\documentclass[10pt,twocolumn,letterpaper]{article}

\usepackage{cvpr}              %

\usepackage{graphicx}
\usepackage{amsmath}
\usepackage{amssymb}
\usepackage{booktabs}

\usepackage{nicefrac}       %
\usepackage{microtype}      %

\usepackage{blindtext}
\usepackage{float}
\usepackage{enumitem}
\usepackage[table]{xcolor}
\usepackage{tabulary,multirow,overpic}
\usepackage{verbatim}
\usepackage{color}
\usepackage{multicol}
\usepackage{dblfloatfix}
\usepackage{pifont}%
\usepackage[accsupp]{axessibility}  %

\newcolumntype{x}[1]{>{\centering\arraybackslash}p{#1pt}}

\newlength\savewidth\newcommand\shline{\noalign{\global\savewidth\arrayrulewidth
		\global\arrayrulewidth 1pt}\hline\noalign{\global\arrayrulewidth\savewidth}}
\newcommand{\tablestyle}[2]{\setlength{\tabcolsep}{#1}\renewcommand{\arraystretch}{#2}\centering\footnotesize}

\definecolor{baselinecolor}{gray}{.92}
\newcommand{\default}[1]{\cellcolor{baselinecolor}{#1}}

\definecolor{demphcolor}{gray}{.2}
\newcommand{\demph}[1]{\textcolor{demphcolor}{#1}}

\definecolor{demphcolor1}{gray}{.5}
\newcommand{\demphs}[1]{\textcolor{demphcolor1}{#1}}

\definecolor{citecolor}{RGB}{34,139,34}
\definecolor{citecolor2}{HTML}{0071bc}
\definecolor{Graylight}{gray}{0.9}
\definecolor{lightred}{RGB}{241,140,142}
\usepackage[pagebackref=true,breaklinks=true,letterpaper=true,colorlinks,
citecolor=citecolor2,bookmarks=false]{hyperref}

\renewcommand{\paragraph}[1]{\vspace{1.25mm}\noindent\textbf{#1}}

\usepackage[capitalize]{cleveref}
\crefname{section}{Sec.}{Secs.}
\Crefname{section}{Section}{Sections}
\Crefname{table}{Table}{Tables}
\crefname{table}{Tab.}{Tabs.}

\def\cvprPaperID{1935} %
\def\confName{CVPR}
\def\confYear{2022}

\newcommand{\app}{\raise.17ex\hbox{$\scriptstyle\sim$}}

\def\x{$\times$}
\newcommand{\figref}[1]{Fig.~\ref{#1}}
\newcommand{\tblref}[1]{Table~\ref{#1}}
\newcommand{\sref}[1]{\S\ref{#1}}

\newcommand{\pool}{\mathcal{P}}

\newcommand{\rbr}[1]{\left(#1\right)}

\newcommand{\boxAP}{AP$^\text{box}$\xspace}
\newcommand{\maskAP}{AP$^\text{mask}$\xspace}
\newcommand{\mvit}{{MViTv2}\xspace}
\newcommand{\expnum}[2]{{#1}\times 10^{{#2}}}
\newcommand{\cmark}{\ding{51}}%
\newcommand{\xmark}{\ding{53}}%

\begin{document}

\title{Multiscale Vision Transformers for Image Classification and Object Detection}

\title{MViTv2: Improved Multiscale Vision Transformers \\ for Classification and Detection  \vspace{-1.0em}}

\author{
	Yanghao Li\textsuperscript{ *, 1} \qquad
	Chao-Yuan Wu\textsuperscript{ *, 1} \qquad
	Haoqi Fan\textsuperscript{ 1} \qquad \\
	Karttikeya Mangalam\textsuperscript{ 1, 2} \qquad
	Bo Xiong\textsuperscript{ 1}\qquad
	Jitendra Malik\textsuperscript{ 1, 2} \qquad
	Christoph Feichtenhofer\textsuperscript{ *, 1}\\
	\small $^{*}$equal technical contribution   \vspace{.5em} \\
	\textsuperscript{1}Facebook AI Research \qquad \qquad \textsuperscript{2}UC Berkeley 
	 \vspace{-1em}
}

\maketitle

\begin{abstract}
\vspace{-2pt}
In this paper, we study Multiscale Vision Transformers (MViTv2) as a unified architecture for image and video classification, as well as object detection. We present an improved version of MViT that incorporates decomposed relative positional embeddings and residual pooling connections. We instantiate this architecture in five sizes and evaluate it for ImageNet classification, COCO detection and Kinetics video recognition where it outperforms prior work. We further compare MViTv2s' pooling attention to window attention mechanisms where it outperforms the latter in accuracy/compute. Without bells-and-whistles, MViTv2 has state-of-the-art performance in 3 domains: 88.8\% accuracy on ImageNet classification, 58.7 \boxAP on COCO object detection as well as 86.1\% on Kinetics-400 video classification. Code and models are available at \url{https://github.com/facebookresearch/mvit}. 
\vspace{-6pt}
\end{abstract}

\section{Introduction}
\label{sec:intro}

Designing architectures for different visual recognition tasks has been historically difficult and the most widely adopted ones have been the ones that combine simplicity and efficacy, \eg \mbox{VGGNet}~\cite{Simonyan2015} and ResNet~\cite{He2015}. More recently Vision Transformers (ViT)~\cite{dosovitskiy2020image} have shown promising performance and are rivaling convolutional neural networks (CNN) and a wide range of modifications have recently been proposed to apply them to different vision tasks~\cite{MViT,arnab2021vivit,beal2020toward,liu2021swin,zheng2021rethinking,strudel2021segmenter,deit,wang2021pyramid}.

While ViT~\cite{dosovitskiy2020image} is popular in image classification, its usage for high-resolution object detection and space-time video understanding tasks remains challenging. 
The density of visual signals poses severe challenges in compute and memory requirements as these scale quadratically in complexity within the self-attention blocks of Transformer-based~\cite{vaswani2017attention} models. 
The community has approached this burden with different strategies: Two popular ones are (1) \textit{local attention} computation within a  window~\cite{liu2021swin} for object detection and (2) \textit{pooling attention} that locally aggregates features before computing self-attention in video tasks~\cite{MViT}.  

The latter fuels Multiscale Vision Transformers (MViT)~\cite{MViT}, an architecture that extends ViT in a simple way: instead of having a fixed resolution throughout the network, it has a feature hierarchy with multiple \emph{stages} starting from high-resolution to low-resolution. MViT is designed for video tasks where it has state-of-the-art performance. %

\begin{figure}[t]
    \centering
    \subfloat[\scriptsize{\textbf{Image classification}}\label{fig:teaser:a}]{%
    \includegraphics[width=0.2767\linewidth]{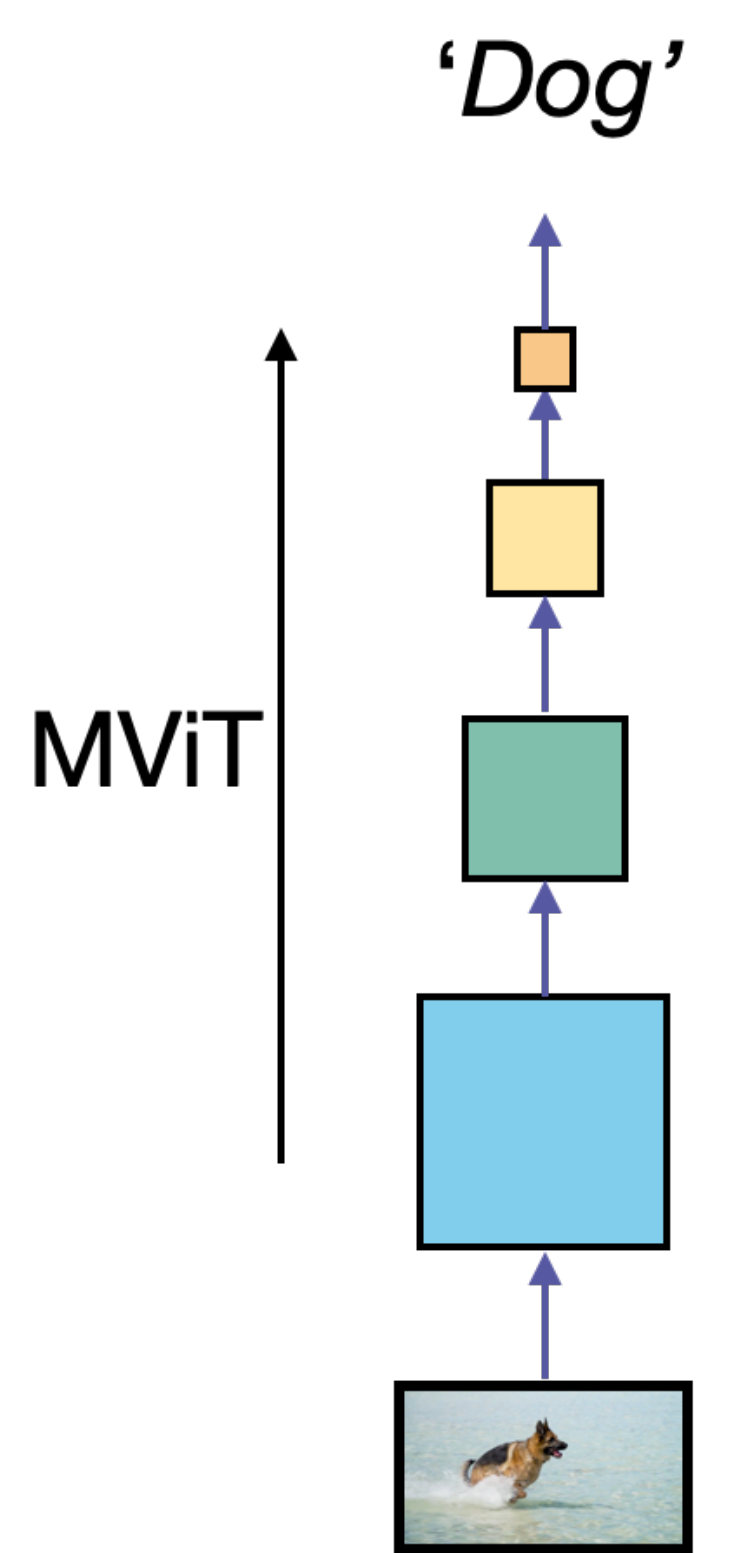}
    }
    \subfloat[\scriptsize{\textbf{Object detection}}\label{fig:teaser:b}]{%
    \includegraphics[width=0.409\linewidth]{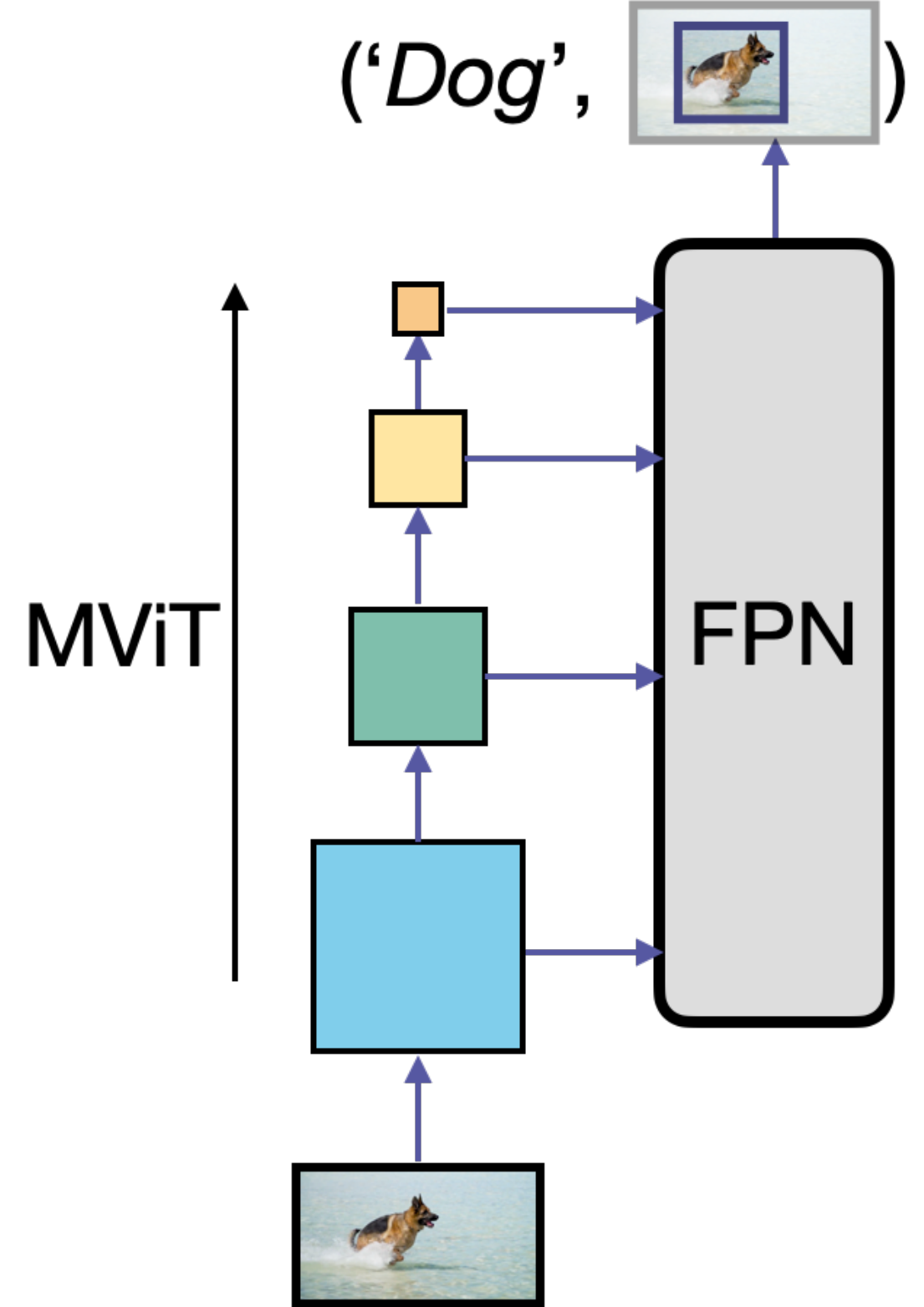}
    }
    \subfloat[\scriptsize{\textbf{Video recognition}}\label{fig:teaser:c}]{%
    \includegraphics[width=0.314\linewidth]{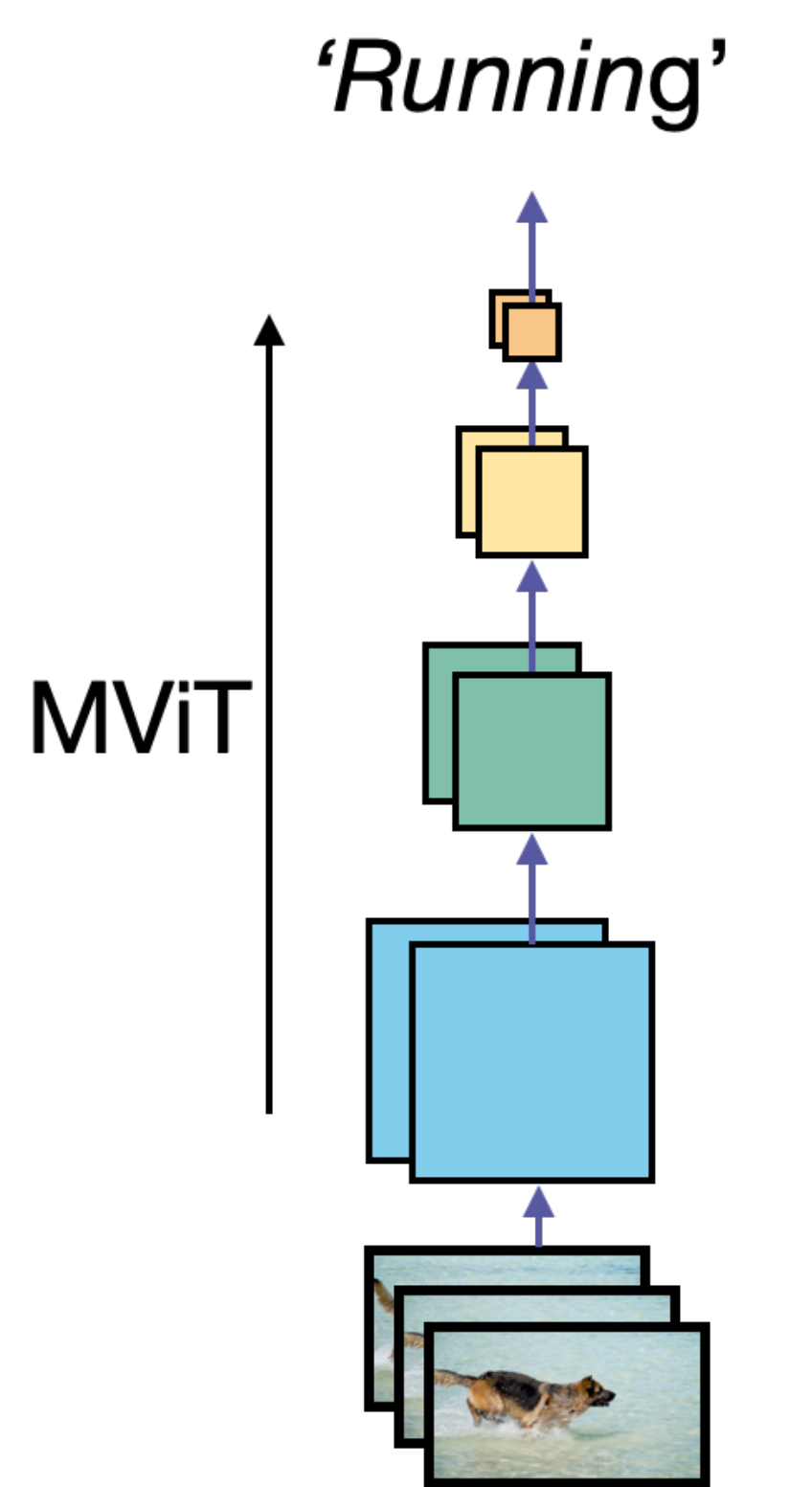}
    }
    \vspace{-4pt}
    \caption{Our \textbf{\mvit} is a multiscale transformer with state-of-the-art performance across three visual recognition tasks. }
     \vspace{-10pt}
    \label{fig:teaser}
\end{figure}

In this paper, we develop two simple technical improvements to further increase its performance and study MViT as a single model family for visual recognition across 3 tasks: image classification, object detection and video classification, in order to understand if it can serve as a \textit{general} vision backbone for spatial as well as spatiotemporal recognition tasks (see~\figref{fig:teaser}). Our empirical study leads to an improved architecture (MViTv2) and encompasses the following:

({\textit{i}}) We create strong baselines that improve pooling attention along two axes: (a) shift-invariant positional embeddings using \textit{decomposed} location distances to inject position information in Transformer blocks; (b) a \emph{residual pooling connection} to compensate the effect of pooling strides in attention computation. 
Our simple-yet-effective upgrades lead to significantly better results. %

({\textit{ii}}) Using the improved structure of MViT, we employ a standard dense prediction framework: \mbox{Mask~R-CNN~\cite{He2017}} with Feature Pyramid Networks (FPN)~\cite{Lin2017} and apply it to object detection and instance segmentation. 

 We study if MViT can process high-resolution visual input by using pooling attention to overcome the computation and memory cost involved. Our experiments suggest that pooling attention is more effective than local window attention mechanisms (\eg Swin~\cite{liu2021swin}). We further develop a simple-yet-effective \emph{Hybrid window attention} scheme that can complement pooling attention for better accuracy/compute tradeoff.

({\textit{iii}}) We instantiate our architecture in five sizes of increasing complexity (width, depth, resolution) and report a practical training recipe for large multiscale transformers. The MViT variants are applied to image classification, object detection and video classification, with minimal modification, to study its purpose as a generic vision architecture.

Experiments reveal that our MViTv2 achieves 88.8\% accuracy for ImageNet-1K classification, with pretraining on ImageNet-21K (and 86.3\% without), as well as {58.7} \boxAP on COCO object detection using only Cascade \mbox{Mask R-CNN}~\cite{cai2018cascade}. For video classification tasks, MViT achieves \textit{unprecedented} accuracies of 86.1\% on \mbox{Kinetics-400}, 87.9\% on Kinetics-600, 79.4\% on Kinetics-700, and 73.3\% on Something-Something-v2.
Our video code will be open-sourced in PyTorchVideo
\footnote{{\scriptsize\url{https://github.com/facebookresearch/pytorchvideo}}}$^\text{,}$\footnote{\url{https://github.com/facebookresearch/SlowFast}}~\cite{fan2021pytorchvideo,fan2020pyslowfast}. 

\section{Related Work}

\paragraph{CNNs} serve as the primary backbones for computer vision tasks, including image recognition~\cite{lecun1989backpropagation,Krizhevsky2012,Simonyan2015,Szegedy2015,He2016a, chen2019drop, tan2019efficientnet, ilija_2020, resnest, dollar2021fast}, object detection~\cite{He2016,Girshick2015,Lin2017,cai2018cascade,Redmon2016,zhou2019objects}  and video recognition~\cite{Simonyan2014,Feichtenhofer2016, Carreira2017, Qiu2017, Li2018, Xie2018,Tran2019, Feichtenhofer2019, Wu2019, girdhar2019video, feichtenhofer2020x3d, Zhou2017, jiang2019stm}. 

\paragraph{Vision transformers} have generated a lot of recent enthusiasm since the work of ViT~\cite{dosovitskiy2020image}, which applies a Transformer architecture on image patches and shows very competitive results on image classification. Since then, different works have been developed to further improve ViT, including efficient training recipes~\cite{deit}, multi-scale transformer structures~\cite{MViT, liu2021swin, wang2021pyramid} and advanced self-attention mechanism design~\cite{MViT,liu2021swin,chu2021twins}.  In this work, we build upon the Multiscale Vision Transformers (MViT) and study it as a \emph{general} backbone for different vision tasks. %

\paragraph{Vision transformers for object detection} tasks \cite{liu2021swin,zhang2021multi,wang2021pyramid,chu2021twins} address the challenge of detection  typically requiring high-resolution inputs and feature maps for accurate object localization. This significantly increases computation complexity due to the quadratic complexity of self-attention operators in transformers~\cite{vaswani2017attention}. Recent works develop technology to alleviate this cost, including %
shifted window attention~\cite{liu2021swin} and Longformer attention~\cite{zhang2021multi}. Meanwhile, pooling attention in MViT is designed to compute self-attention efficiently using a different perspective~\cite{MViT}. In this work, we study MViT for detection and more generally compare pooling attention to local attention mechanisms. %

\vspace{2pt}
\paragraph{Vision transformers for video recognition} have also recently shown strong results, but mostly~\cite{neimark2021video,bertasius2021space,arnab2021vivit,liu2021video} rely on pre-training with large-scale external data (\eg ImageNet-21K~\cite{deng2009imagenet}). MViTv1~\cite{MViT} reports a good training-from-scratch recipe for Transformer-based architectures on Kinetics data~\cite{Kay2017}. In this paper, we use this recipe and improve the MViT architecture with improved pooling attention which is simple yet effective on accuracy; further, we study the (large) effect of ImageNet pre-training for video tasks.

\section{Revisiting Multiscale Vision Transformers} 
The key idea of MViTv1~\cite{MViT} is to construct different \textit{stages} for both low- and high-level visual modeling instead of single-scale blocks in ViT~\cite{dosovitskiy2020image}. 
MViT slowly expands the channel width $D$, while reducing the resolution $L$ (\ie sequence length), from input to output stages of the network.

To perform downsampling within a transformer block, MViT introduces \textit{Pooling Attention}. Concretely, for an input sequence, $X \in \mathbb{R}^{L \times D}$, it applies
linear projections  ${{W}}_Q$, ${{W}}_K$, ${{W}}_V \in \mathbb{R}^{D \times D}$ followed by pooling operators ($\pool$) to query, key and value tensors, respectively:
\begin{align}
    Q = \pool_Q\rbr{X W_Q}, \  K = \pool_K\rbr{X W_K}, \ V = \pool_V\rbr{X W_V},
\end{align}
where the length $\Tilde{L}$ of $Q \in \mathbb{R}^{\Tilde{L} \times D}$ can be reduced by $\pool_Q$ and $K$ and $V$ length can be reduced by $\pool_K$ and $\pool_V$.

Subsequently, pooled self-attention,
\begin{align}\label{eq:z}
    Z := \mathrm{Attn}(Q, K, V) = \mathrm{Softmax}\rbr{QK^\top / \sqrt{D}}V,
\end{align}
computes the output sequence $Z \in \mathbb{R}^{\Tilde{L} \times  D}$ with flexible length $\Tilde{L}$. Note that the downsampling factors $\pool_K$ and $\pool_V$ for key and value tensors can be different from the ones applied to the query sequence, $\pool_Q$.

Pooling attention enables resolution reduction between different stages of MViT by pooling the query tensor $Q$, and to significantly reduce compute and memory complexity by pooling the key, ${K}$, and value, ${V}$, tensors.

\section{Improved Multiscale Vision Transformers}

In this section, we first introduce an empirically powerful upgrade to \emph{pooling attention} (\sref{sec:mvit_v2_improve}). Then we describe how to employ our generic MViT architecture for object detection (\sref{sec:mvit_v2_detection}) and video recognition (\sref{sec:mvit_v2_video}). Finally, \sref{sec:mvit_v2_vairants} shows five concrete instantiations for MViTv2 in increasing complexity.
\subsection{Improved Pooling Attention}\label{sec:mvit_v2_improve}

We start with re-examining two important implications of MViTv2 for potential improvement and introduce techniques %
to understand and address them. %

\definecolor{green1}{HTML}{047513}  %
\definecolor{red1}{HTML}{c9190a}  %
\newcommand{\hlgs}[1]{\textcolor{green1}{#1}}
\newcommand{\hlrs}[1]{\textcolor{red1}{#1}}

\begin{figure}[t!]
	\centering
	\vspace{-15pt}
	\includegraphics[width = 0.9\columnwidth]{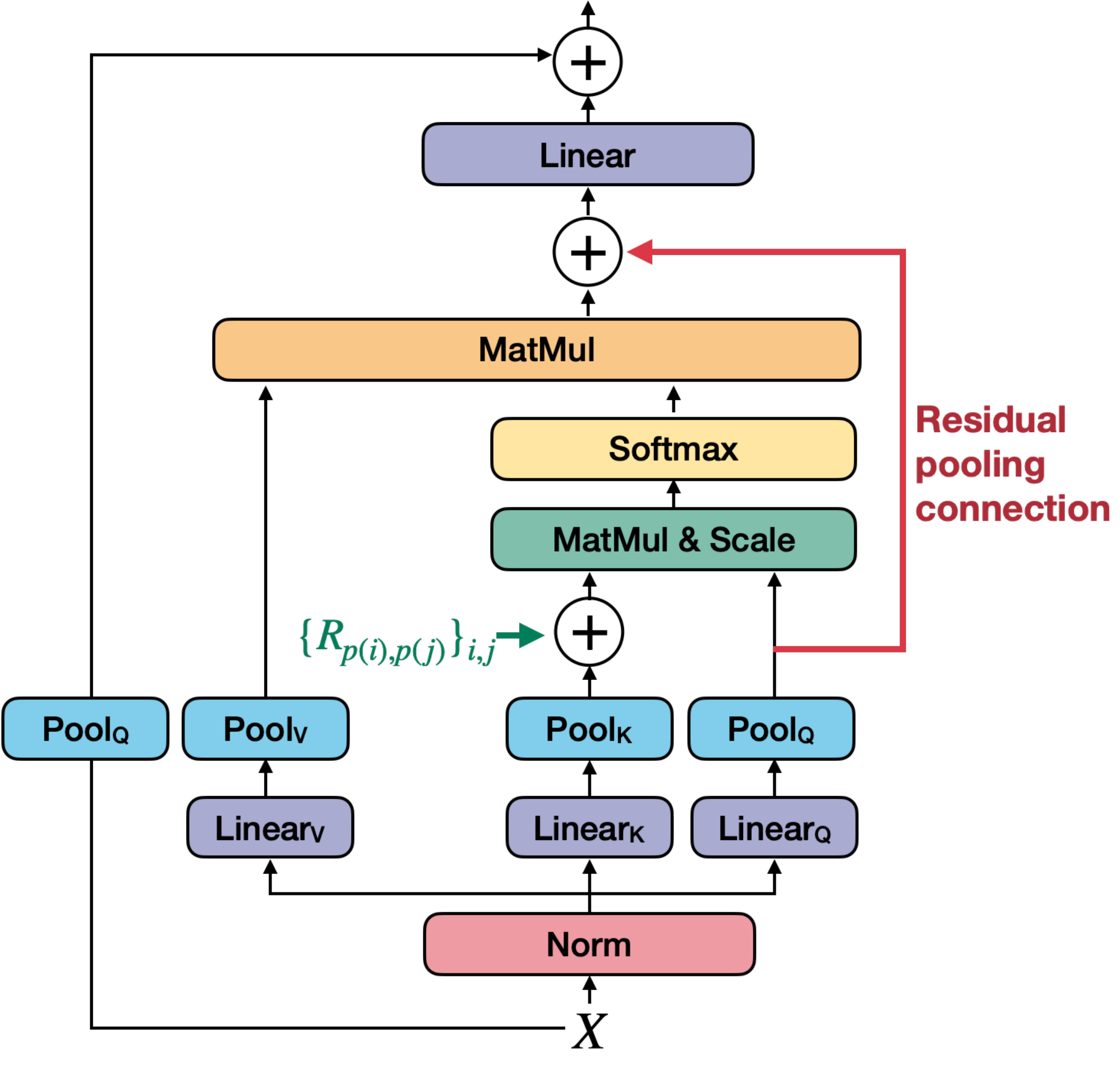}
	\vspace{-13pt}
	\caption{\textbf{The improved Pooling Attention} mechanism that incorporating decomposed relative position embedding, \hlgs{${{R}}_{p(i), p(j)}$}, and \hlrs{residual pooling connection} modules in the attention block.}
    \vspace{-14pt}
	\label{fig:pooling_attn_v2}
\end{figure}

\paragraph{Decomposed relative position embedding.}
While MViT has shown promises in their power to model interactions between tokens, they focus on content, rather than structure.
The space-time structure modeling relies solely on the ``\emph{absolute}" positional embedding to offer location information.
This ignores the fundamental principle of shift-invariance in vision~\cite{lecun1989handwritten}.
Namely, the way MViT models the interaction between two patches will change depending on their \emph{absolute} position in images even if their \emph{relative} positions stay unchanged.
To address this issue, we incorporate relative positional embeddings~\cite{shaw2018self}, which only depend on the relative location distance between tokens into the pooled self-attention computation. %
	
We encode the relative position between the two input elements, $i$ and $j$, into positional embedding ${{R}}_{p(i), p(j)} \in \mathbb{R}^{d}$, where $p(i)$ and $p(j)$ denote the spatial (or spatiotemporal) position of element $i$ and $j$.\footnote{%
Note that $Q$ and ($K$, $V$) can reside in different scales due to potentially different pooling. $p$ maps the index of all of them into a shared scale.} 
The pairwise encoding representation is then embedded into the self-attention module:
\begin{align*}
\mathrm{Attn}(Q, K, V) = \mathrm{Softmax}\rbr{(QK^\top + E^{(\mathrm{rel})}) / \sqrt{d}}V,
\end{align*}
\vspace{-12pt}
\begin{align}
\text{where \quad} E^{(\mathrm{rel})}_{ij} = Q_i \cdot R_{p(i),p(j)}.
\end{align}

However, the number of possible embeddings $R_{p(i),p(j)}$ scale in $\mathcal{O}(TWH)$, which can be expensive to compute.
To reduce complexity, we \textit{decompose} the distance computation between element $i$ and $j$ along the spatiotemporal axes: 
\begin{equation}\label{eq:rel}
{{R}}_{p(i), p(j)} = {{R}}_{h(i),h(j)}^{\mathrm{h}} + {{R}}_{w(i),w(j)}^{\mathrm{w}} + {{R}}_{t(i),t(j)}^{\mathrm{t}},
\end{equation}
where $R^{\mathrm{h}}, R^{\mathrm{w}}, R^{\mathrm{t}}$ are the positional embeddings along the height, width and temporal axes, and $h(i)$, $w(i)$, and $t(i)$ denote the vertical, horizontal, and temporal position of token $i$, respectively. Note that $R^{\mathrm{t}}$ is optional and only required to support temporal dimension in the video case.
In comparison, our decomposed embeddings reduce the number of learned embeddings to $\mathcal{O}(T+W+H)$, which can have a large effect for early-stage, high-resolution feature maps.

\paragraph{Residual pooling connection.} As demonstrated~\cite{MViT}, pooling attention is very effective to reduce the computation complexity and memory requirements in attention blocks. MViTv1 has larger strides on ${K}$ and ${V}$ tensors than the stride of the ${Q}$ tensors which is only downsampled if the resolution of the output sequence changes  across stages. This motivates us to add the residual pooling connection with the (pooled) ${Q}$ tensor to increase information flow and facilitate the training of pooling attention blocks in MViT.

We introduce a new residual pooling connection \emph{inside} the attention blocks as shown in \figref{fig:pooling_attn_v2}. Specifically, we add the pooled query tensor to the output sequence ${Z}$. So Eq.~(\ref{eq:z}) is reformulated as:
\begin{equation}\label{eq:residual_pool}
	Z := \mathrm{Attn}\rbr{Q, K, V} + Q.
\end{equation}
Note that the output sequence ${Z}$ has the same length as the pooled query tensor $Q$. %

The ablations in \sref{sec:exp_video_ablation} and \sref{sec:exp_image_ablation} shows that both the pooling operator ($\pool_Q$) for query $Q$ and the residual path are necessary for the proposed residual pooling connection. This change still enjoys the low-complexity attention computation with large strides in key and value pooling as adding the pooled query sequence in Eq.~\eqref{eq:residual_pool} comes at a low cost.

\subsection{MViT for Object Detection} \label{sec:mvit_v2_detection}

In this section, we describe how to apply the MViT backbone for object detection and instance segmentation tasks.

\begin{figure}[t!]
	\centering
		\vspace{-15pt}
	\includegraphics[width = 1.0\columnwidth]{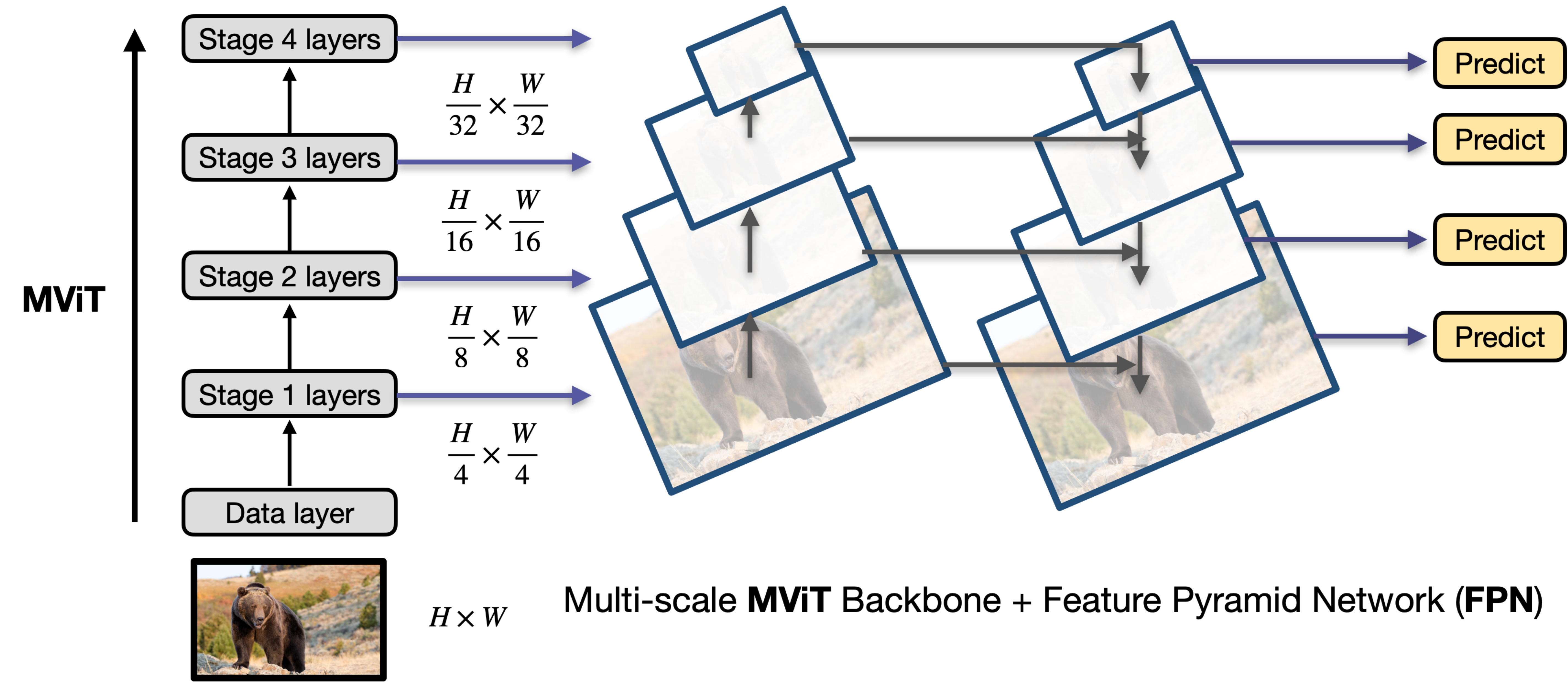}
    \vspace{-15pt}
	\caption{\textbf{MViT backbone used with FPN for object detection. } The multiscale transformer features naturally integrate with standard feature pyramid networks (FPN). %
    }
	\label{fig:mvit_fpn}
	\vspace{-10pt}
\end{figure}

\paragraph{FPN integration.} The hierarchical structure of MViT produces multiscale feature maps in four stages, and therefore naturally integrates into Feature Pyramid Networks (FPN)~\cite{Lin2017} for object detection tasks, as shown in \figref{fig:mvit_fpn}. %
The top-down pyramid with lateral connections in FPN constructs \emph{semantically strong} feature maps for MViT at all scales.
By using FPN with the MViT backbone, we apply it to different detection architectures (\eg Mask R-CNN~\cite{He2017}).

\paragraph{Hybrid window attention.} The self-attention in Transformers has quadratic complexity \wrt the number of tokens. This issue is more exacerbated for object detection as it typically requires high-resolution inputs and feature maps. In this paper, we study two ways to significantly reduce this compute and memory complexity: First, the pooling attention designed in attention blocks of MViT. Second, window attention used as a technique to reduce computation for object detection in Swin~\cite{liu2021swin}.

 Pooling attention and window attention both control the complexity of self-attention by reducing the size of query, key and value tensors when computing self-attention. Their intrinsic nature however is different: Pooling attention pools features by downsampling them via \textit{local aggregation}, but keeps a \textit{global} self-attention computation, while window attention keeps the resolution of tensors but performs self-attention \textit{locally} by dividing the input (patchified tokens) into non-overlapping windows and then only compute local self-attention within each window. The intrinsic difference of the two approaches motivates us to study if they could perform complementary in object detection tasks.

Default window attention only performs local self-attention within windows, thus lacking connections across windows. Different from Swin~\cite{liu2021swin}, which uses shifted windows to mitigate this issue, we propose a simple \emph{Hybrid window attention} (\textbf{Hwin}) design to add cross-window connections. Hwin computes local attention within a window in all but the last blocks of the last three stages that feed into FPN. In this way, the input feature maps to FPN contain global information. The ablation in~\sref{sec:exp_image_ablation} shows that this simple Hwin performs consistently better than Swin~\cite{liu2021swin} on image classification and object detection tasks. Further, we will show that combining pooling attention and Hwin achieves the best performance for object detection. %

\paragraph{Positional embeddings in detection.} Different from ImageNet classification where the input is a crop of fixed resolution (\eg 224$\times$224), object detection typically encompasses inputs of varying size in training. %
For the positional embeddings in MViT (either absolute or relative), we first  initialize the parameters from the ImageNet pre-training weights corresponding to positional embeddings with 224$\times$224 input size and then interpolate them to the respective sizes for object detection training.

\subsection{MViT for Video Recognition} \label{sec:mvit_v2_video}

MViT can be easily adopted for video recognition tasks (\eg the Kinetics dataset) similar to MViTv1~\cite{MViT} as the upgraded modules in \sref{sec:mvit_v2_improve} generalize to the spatiotemporal domain. While MViTv1 only focuses on the training-from-scratch setting on Kinetics, in this work, we also study the (large) effect of pre-training from ImageNet datasets.

\paragraph{Initialization from pre-trained MViT.} Compared to the image-based MViT, there are only three differences for video-based MViT: 1) the projection layer in the \emph{patchification stem} needs to project the input into space-time cubes instead of 2D patches; 2) the pooling operators now pool spatiotemporal feature maps; 3)  relative positional embeddings reference space-time locations.

As the projection layer and pooling operators in 1) and 2) are instantiated by convolutional layers by default~\footnote{
Note that no initialization is needed if using max-pooling variants.}, we use an inflation initialization as for CNNs~\cite{Feichtenhofer2016a,Carreira2017}. Specifically, we initialize the conv filters for the center frame with the weights from  the 2D conv layers in pre-trained models and initialize other weights as zero. For 3), we capitalize on our decomposed relative positional embeddings in Eq.~\ref{eq:rel}, and simply initialize the spatial embeddings from pre-trained weights and the temporal embedding as zero.

\subsection{MViT Architecture Variants} \label{sec:mvit_v2_vairants}

\begin{table}[t!]
    \vspace{-15pt}
    \centering
    \tablestyle{1.6pt}{1.05}
    \begin{tabular}{l|c|c|c|c|r}
        {Model}  & \#Channels  & \#Blocks & \#Heads  & FLOPs   & Param \\
        \shline  
        {MViT}-T & [96-192-384-768] & [1-2-5-2] & [1-2-4-8] & 4.7 & 24 \\
        {MViT}-S & [96-192-384-768] & [1-2-11-2] & [1-2-4-8] & 7.0 & 35 \\
        {MViT}-B & [96-192-384-768] & [2-3-16-3] & [1-2-4-8] & 10.2 & 52 \\
        {MViT}-L & {[144-288-576-1152]} & [2-6-36-4] & [2-4-8-16] & 39.6 & 218   \\
        {MViT}-H & [192-384-768-1536] & [4-8-60-8] & [3-6-12-24] & 120.6 & 667  \\
    \end{tabular}%
    \vspace{-10pt}
    \caption{\textbf{Configuration for MViT variants}. \emph{\#Channels}, \emph{\#Blocks} and \emph{\#Heads} specify the channel width, number of MViT blocks and heads in each block for the four stages, respectively. FLOPs are measured for image classification with 224 $\times$ 224 input. The stage resolutions are [56$^2$, 28$^2$, 14$^2$, 7$^2$]. %
    }
    \label{tab:mvit_v2_variants}
    \vspace{-15pt}
\end{table}

We build several MViT variants with different number of parameters and FLOPs as shown in Table~\ref{tab:mvit_v2_variants}, in order to have a fair comparison with other vision transformer works~\cite{liu2021swin,touvron2020training,wu2021cvt,chen2021crossvit}. Specifically, we design five variants (Tiny, Small, Base, Large and Huge) for MViT by changing the base channel dimension, the number of blocks in each stage and the number of heads in the blocks. Note that we use a smaller number of heads to improve runtime, as more heads lead to slower runtime but have no effect on FLOPs and Parameters. 

Following the pooling attention design in MViT~\cite{MViT}, we employ Key and Value pooling in all pooling attention blocks by default and the pooling stride is set to 4 in the first stage and \textit{adaptively} decays stride w.r.t resolution across stages.

\section{Experiments: Image Recognition}

We conduct experiments on ImageNet classification~\cite{deng2009imagenet} and COCO object detection~\cite{Lin2014}. We first show state-of-the-art comparisons and then perform comprehensive ablations. More results and discussions are in \sref{sec:resultsapp}. 

\subsection{Image Classification on ImageNet-1K}

\paragraph{Settings.} The ImageNet-1K~\cite{deng2009imagenet} (IN-1K) dataset has \app1.28M images in 1000 classes. Our training recipe for \mvit on IN-1K is following MViTv1~\cite{MViT,touvron2020training}. We train all \mvit variants for 300 epochs without using EMA. We also explore pre-training on ImageNet-21K (IN-21K) with \app14.2M images and \app21K classes. See  \sref{app:sec:details} for details.

\begin{table}[t!]
    \vspace{-30pt}
    \centering
    \tablestyle{2.4pt}{1.0}
    \begin{tabular}{l|x{25}x{25}|r|r}
         &   \multicolumn{2}{c|}{Acc}   &   &  \\
        \multicolumn{1}{c|}{model}  & center & resize  & FLOPs (G)   & Param (M) \\
        \shline
    
        {RegNetZ-4GF \cite{dollar2021fast} }    &  \textbf{83.1} && 4.0 &  28 \\ 
        {EfficientNet-B4 $\uparrow$ 380$^2$\cite{tan2019efficientnet} }  &   82.9 && 4.2 & 19 \\
        \hline
        DeiT-S~\cite{touvron2020training}           &  79.8 && 4.6 & 22 \\
        TNT-S~\cite{han2021tnt}                     & 81.5 && 5.2 & 24 \\
        PVTv2-V2~\cite{wang2021pvtv2}              & 82.0 && 4.0 & 25    \\
        CoAtNet-0~\cite{dai2021coatnet}             & 81.6 && 4.2 & 25 \\
        XCiT-S12~\cite{el2021xcit}                  & 82.0 && 4.8 & 26 \\
        Swin-T~\cite{liu2021swin}                   & 81.3 && 4.5 & 29  \\
        CSWin-T~\cite{dong2021cswin}                &  82.7 && 4.3  & 23 \\
        \rowcolor{baselinecolor}
        \textbf{\mvit}-T                          & 82.3 && 4.7 & 24 \\   %
        \shline
        {RegNetY-8GF \cite{ilija_2020} }  &   81.7 && 8.0 &  39 \\
        {EfficientNet-B5 $\uparrow$ 456$^2$\cite{tan2019efficientnet}}  &   \textbf{83.6} && 9.9 & 30\\
        \hline
        
        Twins-B~\cite{chu2021twins}                & 83.2 && 8.6 & 56    \\
        PVTv2-V2-B3~\cite{wang2021pvtv2}              & 83.2 && 6.9 & 45    \\
        Swin-S~\cite{liu2021swin}                  & 83.0 && 8.7 & 50 \\
        CSWin-S~\cite{dong2021cswin}            & \textbf{83.6}& & 6.9 & 35 \\
        MViT-v1-B-16~\cite{MViT} &     83.0 && 7.8 & 37 \\
        \rowcolor{baselinecolor}
        \textbf{\mvit}-S &  \textbf{83.6} && 7.0 & 35 \\   %
        
        \shline
        {RegNetZ-16GF \cite{dollar2021fast} }   &    84.1 && 15.9 &  95 \\
        {EfficientNet-B6 $\uparrow$ 528$^2$\cite{tan2019efficientnet}}  &    84.2 && 19 & 43 \\
        \hline
        DeiT-B \cite{touvron2020training}           &      81.8 && 17.6 & 87 \\
        PVTv2-V2-B5~\cite{wang2021pvtv2}              & 83.8 && 11.8 & 82    \\
        CaiT-S36~\cite{touvron2021going}               & 83.3 && 13.9 & 68 \\
        CoAtNet-2~\cite{dai2021coatnet}             & 84.1 & & 15.7 & 75 \\
        XCiT-M24~\cite{el2021xcit}                  & 82.7 && 16.2 & 84\\
        Swin-B~\cite{liu2021swin}              & 83.3   && 15.4  & 88   \\
        CSWin-B~\cite{dong2021cswin}                                 & 84.2 && 15.0    & 78  \\
        {MViTv1}-B-24~\cite{MViT} &     83.4 && 10.9 & 54 \\
        \rowcolor{baselinecolor}
        \textbf{\mvit}-B &  \textbf{84.4} && 10.2 & 52 \\  %
        \shline
        {EfficientNet-B7 $\uparrow$ 600$^2$\cite{tan2019efficientnet}}  &  84.3 && 37.0 & 66 \\
        NFNet-F1 $\uparrow$ 320$^2$\cite{brock2021high}                         &      84.7 && 35.5 & 133 \\ 
        \hline 
        DeiT-B $\uparrow$ 384$^{2}$ \cite{touvron2020training} &  83.1 &&  55.5 &   87\\
        CvT-32 $\uparrow$ 384$^{2}$~\cite{wu2021cvt}                    && 83.3 & 24.9 & 32    \\
        CaiT-S36$\uparrow$ 384$^{2}$~\cite{touvron2021going}   && 85.0 & 48 & 68 \\
        Swin-B $\uparrow$ 384$^{2}$ \cite{liu2021swin} && 84.2 &  47.0 &   88 \\
        MViT-v1-B-24 $\uparrow$ 320$^{2}$~\cite{MViT} &  {84.8}  &&  32.7 & 73 \\ %
        \rowcolor{baselinecolor}
        \textbf{\mvit}-B $\uparrow$ 384$^{2}$ &  \textbf{85.2} & \textbf{85.6}&	36.7 & 52  \\  %
        \shline
        NFNet-F2 $\uparrow$ 352$^2$\cite{brock2021high}                         &      85.1 && 62.6 & 194 \\ 
        \hline
        CoAtNet-3~\cite{dai2021coatnet}             & 84.5 && 34.7 & 168 \\
       XCiT-M24~\cite{el2021xcit}                  & 82.9 && 36.1 & 189\\
        \rowcolor{baselinecolor}
        \textbf{\mvit}-L  &    \textbf{85.3}  &&  42.1 & 218  \\
        \shline
        NFNet-F4 $\uparrow$ 512$^2$\cite{brock2021high}                         &      85.9 && 215.3 & 316 \\ 
        \hline
        CoAtNet-3~\cite{dai2021coatnet} $\uparrow$ 384$^{2}$   & & 85.8 & 107.4 & 168 \\
        \rowcolor{baselinecolor}
        \textbf{\mvit}-L $\uparrow$ 384$^{2}$  &    \textbf{86.0}  & \textbf{86.3}&  140.2 & 218 \\
    \end{tabular}%
    \vspace{-5pt}
    \caption{\textbf{Comparison to published work on ImageNet-1K}. Input images are 224$\times$224 by default and $\uparrow$ denotes using different sizes. MViT is trained for 300 epochs without any external data or models. We report $\uparrow$ 384$^{2}$ MViT tested with \textit{center} crop or a \textit{resized} view of the original image, to compare to prior work. Full Table in~\ref{tab:sota:in1k_full}
    }
    \label{tab:sota:in1k}
    \vspace{-20pt}
\end{table}

\paragraph{Results using ImageNet-1K.} \tblref{tab:sota:in1k} shows our \mvit and state-of-the-art CNNs and Transformers (without external data or distillation models~\cite{jiang2021token,touvron2021going,yuan2021volo}). The models are split into groups based on computation and compared next. 

Compared to MViTv1~\cite{MViT}, our improved~\mvit has better accuracy with fewer flops and parameters. For example, \mvit-S (83.6\%) improves +0.6\% over MViTv1-B-16 (83.0\%) with ~10\% fewer flops. On the base model size, \mvit-B (84.4\%) improves +1.0\% over MViTv1-B-24 (83.4\%) while even being lighter.  
This shows clear effectiveness of the \mvit improvements in \sref{sec:mvit_v2_improve}.

Our \mvit outperforms other Transformers, including DeiT~\cite{touvron2020training} and Swin~\cite{liu2021swin}, %
especially when scaling up models. For example, \mvit-B achieves 84.4\% top-1 accuracy, surpassing DeiT-B and Swin-B by 2.6\% and 1.1\% respectively. Note that \mvit-B has over 33\% fewer flops and parameters comparing DeiT-B and Swin-B. The trend is similar with 384$\times$384 input and \mvit-B has further +0.8\% gain from the high-resolution fine-tuning under center crop testing.

In addition to  \textit{center} crop testing (with a 224$/$256=0.875 crop ratio), we report a testing protocol that has been adopted recently in the community~\cite{touvron2021going,liu2021swin,wu2021cvt}: This protocol takes a \textit{full}-sized crop of the (resized) original validation images. We observe that \textit{full} crop testing can increase our \mvit-L $\uparrow$~384$^{2}$ from 86.0 to \textbf{86.3}\%, which is the highest accuracy on IN-1K to date (without external data or distillation models).

\paragraph{Results using ImageNet-21K.} Results for using the large-scale IN-21K pre-training are shown in \tblref{tab:sota:in1k-ft}. The IN-21K data adds +2.2\% accuracy to \mvit-L. 

Compared to other Transformers, \mvit-L achieves better results than \mbox{Swin-L} (+1.2\%). We lastly finetune \mvit-L with 384$^2$ input to directly compare to prior models of size L: \mvit-L achieves 88.4\%, outperforming other large models. We further train a huge \mvit-H  with accuracy 88.0\%, {88.6}\% and \textbf{88.8}\% at 224$^2$, 384$^2$ and 512$^2$ resolution. %

\begin{table}[t!]
    	\vspace{-30pt}
        \centering
        \tablestyle{2.4pt}{1.01}
            \begin{tabular}{l|cc|r|r}
                 &   \multicolumn{2}{c|}{Acc}   &   &  \\

                \multicolumn{1}{c|}{model}  & center & resize  & FLOPs (G)   & Param (M) \\
                \shline
    		Swin-L \cite{liu2021swin} &     86.3 & & 34.5 & 197 \\ 		%
           
            \rowcolor{baselinecolor}
    		\textbf{\mvit}-L &  {87.5} & &  42.1 & 218 \\
    		\rowcolor{baselinecolor}
    		\textbf{\mvit}-H &  \textbf{88.0} & &  120.6 & 667 \\ %
            \shline
            ViT-L/16 $\uparrow$ 384$^{2}$\cite{dosovitskiy2020image}  &     85.2 & & 190.7 & 307 \\
            ViL-B-RPB $\uparrow$ 384$^{2}$~\cite{zhang2021multi}   &  86.2  & & 43.7  & 56 \\
            Swin-L $\uparrow$ 384$^{2}$\cite{liu2021swin} &  &   87.3 & 103.9 & 197 \\ 	
            CSwin-L $\uparrow$ 384$^{2}$\cite{dong2021cswin} & &     87.5 & 96.8 & 173 \\      
            CvT-W24 $\uparrow$ 384$^{2}$\cite{wu2021cvt} &  &   87.6 & 193.2 & 277 \\ 
            CoAtNet-4~\cite{dai2021coatnet} $\uparrow$ 512$^{2}$   & & 88.4 & 360.9 & 275 \\ \hline
            \rowcolor{baselinecolor}
        	\textbf{\mvit}-L $\uparrow$ 384$^{2}$  &  {88.2} & 88.4 &  140.7  &   218 \\ 
        	\rowcolor{baselinecolor}
            \textbf{\mvit}-H $\uparrow$ 384$^{2}$  &  88.3 & {88.6} & 388.5  &   667 \\   
           	\rowcolor{baselinecolor}
            \textbf{\mvit}-H $\uparrow$ 512$^{2}$  &  {88.3} & \textbf{88.8} & 763.5  &   667 \\ %

            \end{tabular}%
        \vspace{-10pt}
        \caption{\textbf{ImageNet-1K fine-tunning} results using \textbf{IN-21K} data. Fine-tuning is with 224$^2$ input size (default) or with $\uparrow$ 384$^{2}$ size. Center denotes testing with a center crop, while resize is scaling the full image to the inference resolution (including more context). 
        }
        \label{tab:sota:in1k-ft}
        \vspace{-15pt}
    \end{table}

\subsection{Object Detection on COCO}
 \vspace{-2pt}

     \begin{table*}[b!]
    \vspace*{-10pt}
     \captionsetup{position=top}
     \subfloat[\textbf{ImageNet-1K classification}\label{tab:ablation:attention_IN}]{%
         \tablestyle{1.0pt}{1.05}
         \begin{tabular}{l|l|x{33}cc}
            variant & attention & Acc & FLOPs (G) & Mem (G) \\
            \shline
            \multirow{5}{*}{ViT-B} & full & 82.0	& 17.5&	12.4\\
            & fixed win & 80.0	&17.0	&9.7\\
             & Swin~\cite{liu2021swin}  & 80.4	&17.0&	9.7 \\
            & Hwin  & \textbf{82.1}	&17.1	&10.4\\
            & pooling  & 81.9	&\textbf{10.9}	&\textbf{8.3}\\
            \hline
            
            \multirow{4}{*}{\mvit-S} & \default{pooling} & \default{\textbf{83.6}} & \default{7.0} & \default{6.8}\\
            & pooling (stride=8)  & 83.2 & \textbf{6.3} & \textbf{5.5}\\
            & pooling  + Swin~\cite{liu2021swin}  & 82.8 & 6.4 & 6.0\\   
            & pooling + Hwin  & 83.0	& 6.5 & 6.2 \\
            \multicolumn{5}{c}{~}\\
        \end{tabular}
     }\hfill  %
     \subfloat[\textbf{Mask R-CNN on COCO detection}\label{tab:ablation:attention_coco}]{%
         \tablestyle{1.0pt}{1.05}
         \begin{tabular}{l|l|x{33}ccccc}
            \multirow{1}{*}{variant} & \multirow{1}{*}{attention} & \multirow{1}{*}{AP$^\text{box}$}  & Train(iter/s)  & Test(im/s)  & Mem(G) \\ %
            \shline
            \multirow{6}{*}{ViT-B} & full   & 46.6	 &2.3	&4.6	& 24.7 \\
             & fixed win                      & 43.4	 & \textbf{3.3}	&7.8	&5.6 \\
             & Swin~\cite{liu2021swin}          & 45.1   &3.1	&7.5	&5.7 \\
             & Hwin                              & 46.1	& 3.1	&6.8	&11.0 \\
             & pooling                           & \textbf{47.2}	& 2.9	&7.9	&8.8 \\
             & pooling + Hwin                    & 46.9	& 3.1	&\textbf{8.8}	&\textbf{5.5} \\
            \hline
            \multirow{4}{*}{\mvit-S} & pooling   & \textbf{50.8}  & 1.5 & 4.2 & 19.5 \\
             & pooling (stride=8)    & 50.0 & 2.5 & 8.3 & 7.8 \\
            & pooling  + Swin~\cite{liu2021swin} & 48.9 & 2.6 & 9.2 & \textbf{4.9} \\
            & \default{pooling + Hwin } & \default{49.9} &	 \default{\textbf{2.7}} & \default{\textbf{9.4}} & \default{5.2} \\
        \end{tabular}
     }\hfill  %
     \caption{\textbf{Comparison of attention mechanisms on ImageNet and COCO} using ViT-B and \mvit-S backbones. \emph{fixed win}: non-overlapping window-attention in all Transformer blocks. \emph{Swin}: shifted window attention~\cite{liu2021swin}. \emph{Hwin}: our Hybrid window attention. \emph{Pooling}: our pooling attention, the K, V pooling stride is 2 (ViT-B) and 4 on the first stage of \mvit, or \emph{pooling (stride=8)}. Accuracy, FLOPs and peak training memory are measured on IN-1K. For COCO, we report \boxAP, average training iterations per-second, average testing frames per-second and peak training memory, which are measured in Detectron2~\cite{wu2019detectron2} with 8 V100 GPUs under the same settings. Default is in gray. %
     \label{tab:ablation:attention}
           \vspace*{-20pt}
     } 
 \end{table*}     

\begin{table}[t]
      \vspace{-10pt}
          \captionsetup{position=top}
     \subfloat[\textbf{Mask R-CNN}\label{tab:sota_coco:mask}]{%
     \centering
         \tablestyle{1.0pt}{1.04}
         \hspace{-12pt}
         \begin{tabular}{l|ccc|ccc|cc}
             model & {AP$^\text{box}$} & \demphs{AP$^\text{box}_\text{50}$} & \demphs{AP$^\text{box}_\text{75}$} & {AP$^\text{mask}$} & \demphs{AP$^\text{mask}_\text{50}$} & \demphs{AP$^\text{mask}_\text{75}$} & FLOPs   & Param \\
              \shline
              Res50~\cite{He2016}                     & 41.0 & \demphs{61.7} & \demphs{44.9} & 37.1 & \demphs{58.4} & \demphs{40.1}  & 260 & 44\\
              PVT-S~\cite{wang2021pyramid}            & 43.0 & \demphs{65.3} & \demphs{46.9} & 39.9 & \demphs{62.5} & \demphs{42.8}  & 245 & 44 \\
              Swin-T~\cite{liu2021swin}                & 46.0 & \demphs{68.2} & \demphs{50.2} & 41.6 & \demphs{65.1} & \demphs{44.8}  & 264 & 48 \\
              ViL-S-RPB~\cite{zhang2021multi}         & 47.1 & \demphs{68.7} & \demphs{51.5} & 42.7 & \demphs{65.9} & \demphs{46.2} & 277 & 45 \\ 
              MViTv1-T~\cite{MViT}                    & 45.9 & \demphs{68.7} & \demphs{50.5} & 42.1 & \demphs{66.0} & \demphs{45.4} &  326 & 46 \\
              \rowcolor{baselinecolor} 					
              \textbf{\mvit}-T & \textbf{48.2}	& \demphs{\textbf{70.9}}	& \demphs{\textbf{53.3}}	& {\textbf{43.8}}	& \demphs{\textbf{67.9}}	& \demphs{\textbf{47.2}} & 279 & 44  \\ %
              \shline
              Res101~\cite{He2016}                & 42.8 & \demphs{63.2} & \demphs{47.1} & 38.5 & \demphs{60.1} & \demphs{41.3}  & 336 & 63\\
              PVT-M~\cite{wang2021pyramid}        & 44.2 & \demphs{66.0} & \demphs{48.2} & 40.5 & \demphs{63.1} & \demphs{43.5}  & 302 & 64 \\
              Swin-S~\cite{liu2021swin}           & 48.5 & \demphs{70.2} & \demphs{53.5} & 43.3 & \demphs{67.3} & \demphs{46.6}  & 354 & 69\\
              ViL-M-RPB~\cite{zhang2021multi}     & 48.9 & \demphs{70.3} & \demphs{54.0} & 44.2 & \demphs{67.9} & \demphs{47.7} & 352 & 60 \\ 
              MViTv1-S~\cite{MViT}               & 47.6 & \demphs{	70.0} & \demphs{52.2} & 43.4 & \demphs{67.3} & \demphs{46.9} & 373 & 57 \\  				
              \rowcolor{baselinecolor}
              \textbf{\mvit}-S & \textbf{49.9}	& \demphs{\textbf{72.0}}	&\demphs{\textbf{55.0}}	&\textbf{45.1}	&\demphs{\textbf{69.5}}	&\demphs{\textbf{48.5}}   & 326 & 54 \\
              \shline
              X101-64~\cite{Xie2017}           & 44.4 & \demphs{64.9} & \demphs{48.8} & 39.7 & \demphs{61.9} & \demphs{42.6}  & 493 & 101\\
              PVT-L\cite{wang2021pyramid}      & 44.5 & \demphs{66.0} & \demphs{48.3} & 40.7 & \demphs{63.4} & \demphs{43.7}  & 364 & 81 \\
              Swin-B~\cite{liu2021swin}        & 48.5 & \demphs{69.8} & \demphs{53.2} & 43.4 & \demphs{66.8} & \demphs{46.9}  & 496 & 107\\
              ViL-B-RPB~\cite{zhang2021multi}          & 49.6 & \demphs{70.7} & \demphs{54.6} & 44.5 & \demphs{68.3} & \demphs{48.0} & 384 & 76 \\
              MViTv1-B~\cite{MViT}                     & 48.8 & \demphs{71.2} & \demphs{53.5} & 44.2 & \demphs{68.4} & \demphs{47.6} & 438 & 73        \\ 					
              \rowcolor{baselinecolor}
              \textbf{\mvit}-B & \textbf{51.0}	& \demphs{\textbf{72.7}}	&\demphs{\textbf{56.3}}	&\textbf{45.7}	&\demphs{\textbf{69.9}}	&\demphs{\textbf{49.6}}  & 392 & 71\\  %
              \shline
              \rowcolor{baselinecolor}
              \textbf{\mvit}-L & 51.8	&\demphs{72.8}	&\demphs{56.8}	&46.2	&\demphs{70.4}	&\demphs{50.0} & 1097 & 238 \\
              \rowcolor{baselinecolor}
              \textbf{\mvit}-L$\dagger$ & \textbf{52.7}	&\demphs{\textbf{73.7}}	&\demphs{\textbf{57.6}}	&\textbf{46.8}	&\demphs{\textbf{71.4}}	&\demphs{\textbf{50.8}} & 1097 & 238  \\  
          \end{tabular}
     }\hfill  %
     \vspace{2mm}
     \subfloat[\textbf{Cascade Mask R-CNN}\label{tab:sota_coco:cascade}]{%
     \centering
         \tablestyle{1.0pt}{1.04}
         \hspace{-10pt}
         \begin{tabular}{l|ccc|ccc|cc}
             model & {AP$^\text{box}$} & \demphs{AP$^\text{box}_\text{50}$} & \demphs{AP$^\text{box}_\text{75}$} & {AP$^\text{mask}$} & \demphs{AP$^\text{mask}_\text{50}$} & \demphs{AP$^\text{mask}_\text{75}$} & FLOPs   & Param \\
              \shline
              R50~\cite{He2016}               & 46.3 & \demphs{64.3} & \demphs{50.5} & 40.1 & \demphs{61.7} & \demphs{43.4}  & 739 & 82 \\
              Swin-T~\cite{liu2021swin}       & 50.5 & \demphs{69.3} & \demphs{54.9} & 43.7 & \demphs{66.6} & \demphs{47.1}  & 745 & 86\\
              \rowcolor{baselinecolor}
              \textbf{\mvit}-T &  \textbf{52.2}	&\demphs{\textbf{71.1}}	&\demphs{56.6}	&\textbf{45.0}	&\demphs{\textbf{68.3}}	&\demphs{\textbf{48.9}}& 701 &76\\  %
              \shline
              X101-32~\cite{Xie2017}       & 48.1 & \demphs{66.5} & \demphs{52.4} & 41.6 & \demphs{63.9} & \demphs{45.2}   & 819 & 101\\
              Swin-S~\cite{liu2021swin}    & 51.8 & \demphs{70.4} & \demphs{56.3} & 44.7 & \demphs{67.9} & \demphs{48.5}  & 838 & 107 \\
              \rowcolor{baselinecolor}
              \textbf{\mvit}-S & \textbf{53.2}	&\demphs{\textbf{72.4}}	&\demphs{58.0}	&\textbf{46.0}	&\demphs{\textbf{69.6}}	&\demphs{50.1} & 748 & 87\\
              \shline
              X101-64~\cite{Xie2017}~~~~~        & 48.3 & \demphs{66.4} & \demphs{52.3} & 41.7 & \demphs{64.0} & \demphs{45.1}  & 972 & 140\\
              Swin-B~\cite{liu2021swin}         & 51.9 & \demphs{70.9} & \demphs{56.5} & 45.0 & \demphs{68.4} & \demphs{48.7}  & 982 & 145\\
              \rowcolor{baselinecolor}
              \textbf{\mvit}-B &        {54.1}	&\demphs{{72.9}}	&\demphs{{58.5}}	&{46.8}	&\demphs{{70.6}}	&\demphs{{50.8}} & 814 &103 \\
               \textbf{\mvit}-B$\dagger$ &       \textbf{54.9}	&\demphs{\textbf{73.8}} &\demphs{\textbf{59.8}}	&\textbf{47.4} &\demphs{\textbf{71.5}}	&\demphs{\textbf{51.6}} & 814 &103 \\
              \shline
              \rowcolor{baselinecolor}
              \textbf{\mvit}-L              &  54.3	       &     \demphs{73.1} &	\demphs{59.1} & 47.1	&\demphs{70.8}	&\demphs{51.7} & 1519 & 270 \\
              \rowcolor{baselinecolor}
              \textbf{\mvit}-L$\dagger\dagger$     & {55.8}	        & \demphs{{74.3}}    & \demphs{{60.9}}	&\textbf{48.3}	&\demphs{{71.9}}	&\demphs{{53.2}}  & 1519 & 270 \\
            \rowcolor{baselinecolor}
            \textbf{\mvit}-H$\dagger\dagger$    & \textbf{56.1}	        & \demphs{\textbf{74.6}}    & \demphs{\textbf{61.0}}	&\textbf{48.5}	&\demphs{\textbf{72.4}}	&\demphs{\textbf{53.2}}  & 3084 & 718 \\
            \hline
            \rowcolor{baselinecolor}
            \textbf{\mvit}-L$\dagger\dagger^*$     & \textbf{58.7}	        & \demphs{\textbf{76.7}}    & \demphs{\textbf{64.3}}	&\textbf{50.5}	&\demphs{\textbf{74.2}}	&\demphs{\textbf{55.9}}  & - & 270 \\
          \end{tabular}
     }\hfill  %
     \vspace{-8pt}
     \caption{\textbf{Results on COCO object detection} with \protect\subref{tab:sota_coco:mask} Mask R-CNN~\cite{He2017} and \protect\subref{tab:sota_coco:cascade} Cascade Mask R-CNN~\cite{cai2018cascade}. $\dagger$ indicates that the model is initialized from IN-21K pre-training. %
     $\dagger\dagger$ denotes using a stronger \emph{large-scale jittering} training~\cite{ghiasi2021simple} and longer schedule (50 epochs) with IN-21K pre-training %
     $^*$~indicates using SoftNMS and multiscale testing.
     FLOPs / Params are in Giga ($10^9$) / Mega ($10^6$). %
     }
     \vspace{-35pt}
     \label{tab:sota_coco}
    
 \end{table}

\paragraph{Settings.} We conduct object detection experiments on the MS-COCO dataset~\cite{Lin2014}. All the models are trained on 118K training images and evaluated on the 5K validation images. We use standard Mask R-CNN~\cite{He2017} and Cascade Mask R-CNN~\cite{cai2018cascade} detection frameworks implemented in Detectron2~\cite{wu2019detectron2}. For a fair comparison, we follow the same recipe as in Swin~\cite{liu2021swin}. Specifically, we pre-train the backbones on IN and fine-tune on COCO using a 3\x schedule (36 epochs) by default. Detailed training recipes are in \sref{app:sec:coco_details}.

For \mvit, we take the backbone pre-trained from IN and add our Hybrid window attention (Hwin) by default. The window sizes are set as [56, 28, 14, 7] for the four stages, which is consistent with the self-attention size used in IN pre-training which takes 224$\times$224 as input.

\paragraph{Main results.} Table~\ref{tab:sota_coco:mask} shows the results on COCO using \mbox{Mask~R-CNN}. Our \mvit surpasses CNN (\ie ResNet~\cite{He2016} and ResNeXt~\cite{Xie2017}) and Transformer backbones (\eg Swin~\cite{liu2021swin}, ViL~\cite{zhang2021multi} and MViTv1~\cite{MViT}\footnote{We adapt MViTv1~\cite{MViT} as a detection baseline combined with Hwin.}). %
\textit{E.g.}, \mbox{\mvit-B} outperforms Swin-B by +2.5/+2.3 in \boxAP/\maskAP, with lower compute and smaller model size. 
When scaling up, our deeper \mvit-L improves over \mvit-B by +0.8 \boxAP and using IN-21K pre-training further adds +0.9 to achieve 52.7 \boxAP with Mask R-CNN and a standard 3\x schedule.  %

In Table~\ref{tab:sota_coco:cascade} we observe a similar trend among backbones for Cascade Mask~R-CNN~\cite{cai2018cascade} which lifts Mask R-CNN accuracy (\ref{tab:sota_coco:mask}).
We also ablate the use of a longer training schedule with large-scale jitter that boosts our \boxAP to {55.8}. \mvit-H increases this to \textbf{56.1} \boxAP and \textbf{48.5} \maskAP. %

We further adopt two inference strategies (SoftNMS~\cite{bodla2017soft} and multi-scale testing) on \mvit-L with Cascade Mask R-CNN for system-level comparison (See Table~\sref{app:tab:coco_system_results}). They boosts our \boxAP to \textbf{58.7}, which is already better than the best results from Swin (58.0 \boxAP), even \mvit does not use the improved HTC++ detector~\cite{liu2021swin} yet.

\subsection{Ablations on ImageNet and COCO}\label{sec:exp_image_ablation}
\vspace{-5pt}

\paragraph{Different self-attention mechanism.} We first study our pooling attention and Hwin self-attention mechanism in \mvit by comparing with different self-attention mechanisms on ImageNet and COCO. For a fair comparison, we conduct the analysis on both ViT-B and \mvit-S networks.

In \tblref{tab:ablation:attention_IN} we compare different attention schemes on IN-1K. We compare 5 attention mechanisms: global (full), windowed, Shifted window (Swin), our Hybrid window (Hwin) and pooling. 
We observe the following:

(\textit{i}) For ViT-B based models, \emph{default win} reduces both FLOPs and Memory usage while the top-1 accuracy also drops by 2.0\% due to the missing cross-window connection. Swin~\cite{liu2021swin} attention can recover 0.4\% over \emph{default win}. While our Hybrid window (Hwin) attention  fully recovers the performance and outperforms Swin attention by \textbf{+1.7}\%. Finally, \emph{pooling attention} achieves the best accuracy/computation trade-off by getting similar accuracy for ViT-B with significant  compute reduction (\app38\% fewer FLOPs). 

(\textit{ii}) For \mvit-S, pooling attention is used by default. We study if adding local window attention can improve MViT. %
We observe that adding Swin or Hwin both can reduce the model complexity with slight performance decay. However, directly increasing the pooling stride (from 4 to 8) achieves the best accuracy/compute tradeoff. 

\tblref{tab:ablation:attention_coco} shows the comparison of attention mechanisms on COCO: (\textit{i}) For ViT-B based models, \emph{pooling} and \emph{pooling + Hwin} achieves even better results (+0.6/0.3 \boxAP) than \emph{standard} full attention with \app2\x~test speedup. (\textit{ii}) For \mvit-S, directly increasing the pooling stride (from 4 to 8) achieves better accuracy/computation tradeoff than adding \emph{Swin}.
This result suggests that simple pooling attention can be a strong baseline for object detection. Finally, combining our \emph{pooling} and \emph{Hwin} achieves the best tradeoff.

\definecolor{green}{HTML}{39b54a}  %
\definecolor{red}{HTML}{ea4335}  %
\newcommand{\hlg}[1]{\textcolor{green}{#1}}
\newcommand{\hlr}[1]{\textcolor{red}{#1}}
\newcolumntype{z}{>{\centering\arraybackslash}p{31pt}}
\newcolumntype{y}{>{\raggedright\arraybackslash}p{68pt}}
\newcolumntype{d}[1]{>{\raggedright\arraybackslash}p{#1pt}}
\newcolumntype{b}[1]{>{\raggedleft\arraybackslash}p{#1pt}}

\newcommand{\better}[2]{\tablestyle{1pt}{1}
\begin{tabular}{b{16}d{16}}
{#1} &
{\fontsize{7pt}{1em}\selectfont \hlg{$\uparrow$#2}}
\end{tabular}
}
\newcommand{\worse}[2]{\tablestyle{1pt}{1}
\begin{tabular}{b{16}d{16}}
{#1} &
{\fontsize{7pt}{1em}\selectfont \hlr{$\downarrow$#2}}
\end{tabular}
}
\newcommand{\void}[1]{\tablestyle{1pt}{1}
\begin{tabular}{b{16}d{16}}
{#1} &
~
\end{tabular}
}
\begin{table}[h!]
    \vspace{-10pt}
    \centering
    \tablestyle{1.4pt}{1.02}
    \begin{tabular}{l|c|cccc}
        \multirow{2}{*}{positional embeddings} & \multicolumn{1}{c|}{IN-1K} & \multicolumn{4}{c}{COCO} \\
        & \multirow{1}{*}{Acc}  & \multirow{1}{*}{AP$^\text{box}$} &  Train(iter/s)  & Test(im/s)  & Mem(G)  \\
        \shline 
        (\emph{1}) no pos.             & 83.3 & 49.2 & \textbf{3.1}	&\textbf{10.3}	& \textbf{5.0}\\
        (\emph{2}) abs.~pos.           & 83.5 & 49.3	&  \textbf{3.1}	& 10.1	& \textbf{5.0} \\
               (\emph{3}) joint rel.~pos.     & 83.6 & \textbf{49.9} & \worse{0.7}{\textbf{4.4}\x} & \worse{3.4}{\textbf{3}\x} & 15.3 \\
        \rowcolor{baselinecolor}
        (\emph{4}) \default{\textit{decomposed} rel.~pos.}  & 83.6 &  \textbf{49.9}	&  2.7	& 9.4	& 5.2\\
        (\emph{5}) abs.~+ dec.~rel.~pos.      & \textbf{83.7} & 49.8	& 2.7	& 9.5	& 5.2\\
    \end{tabular}
    \vspace{-10pt}
    \caption{\textbf{Ablation of positional embeddings} on \mvit-S.~~~~~~~~~ %
    }
    \label{tab:ablation:pos}
    \vspace{-13pt}
\end{table}

\paragraph{Positional embeddings.} \tblref{tab:ablation:pos} compares different positional embeddings. We observe that: (\textit{i}) Comparing (\emph{2}) to (\emph{1}), \emph{absolute position} only slightly improves over \emph{no pos.}. This is because the pooling operators (instantiated by conv layers) already model positional information. 
(\textit{ii}) Comparing (\emph{3}, \emph{4}) and (\emph{1}, \emph{2}), relative positions can bring performance gain by introducing shift-invariance priors to pooling attention. Finally, our \emph{decomposed relative position} embedding train \textbf{3.9}\x~\textit{faster} than \emph{joint relative position} on COCO.

\begin{table}[h!]
    \centering
    \vspace{-10pt}
    \tablestyle{1.4pt}{1.02}
    \begin{tabular}{l|c|cccc}
        \multirow{2}{*}{residual pooling} & \multicolumn{1}{c|}{IN-1K} & \multicolumn{4}{c}{COCO}  \\
        & Acc  & AP$^\text{box}$ &  Train(iter/s)  & Test(im/s)  & Mem(G)\\
        \shline
        (\emph{1}) w/o                                     & 83.3 & 48.5	& \textbf{3.0}    & \textbf{10.0} & \textbf{4.7}	\\
        (\emph{2}) residual                                & \textbf{83.6} & 49.3   &	2.9	& 9.8 & \textbf{4.7}\\
        \rowcolor{baselinecolor}
        (\emph{3}) \default{{\scriptsize full Q pooling + residual}}     & \textbf{83.6} & \textbf{49.9}	&  2.7	& 9.4	& 5.2\\
        (\emph{4}) full Q pooling                          & 83.1 & 48.5   &	 2.8 &	9.5 & 5.1\\

    \end{tabular}
    \vspace{-10pt}
   \caption{\textbf{Ablation of residual pooling connections} on \mvit-S.%
    }
    \label{tab:ablation:residual}
    \vspace{-15pt}
\end{table}

\paragraph{Residual pooling connection.} \tblref{tab:ablation:residual} studies the importance of our residual pooling connection. We see that simply adding the residual path (\emph{2}) can improves results on both IN-1K (+0.3\%) and COCO (+0.8 for \boxAP) with \textit{negligible} cost. (\emph{3}) Using residual pooling and also adding Q pooling to all other layers (with stride=1) leads to a significant boost, especially on COCO (+1.4 \boxAP). This suggests both Q pooling blocks and residual paths are necessary in \mvit. (\emph{4}) just adding (without residual) more Q pooling layers with stride=1 does not help and even decays (\emph{4}) \vs (\emph{1}).

\begin{table}[t]
    \centering
    \small
    \vspace{-10pt}
    \tablestyle{1.1pt}{1.05}
    \begin{tabular}{l|cc|cccc}

        \multirow{2}{*}{model} & \multicolumn{2}{c|}{IN-1K} & \multicolumn{4}{c}{COCO} \\
        & \multirow{1}{*}{Acc}  & Test (im/s)~ & \multirow{1}{*}{AP$^\text{box}$} &  Train(iter/s)  &  Test(im/s)  & Mem(G) \\
        \shline
        Swin-B~\cite{liu2021swin} & 83.3 & 276 &  48.5 	& 2.5 & \textbf{9.4} &	6.3 \\ %
        \hline
        \rowcolor{baselinecolor}
        \mvit-S & 83.6 & \textbf{341} & 49.9 &  \textbf{2.7}	& \textbf{9.4} & \textbf{5.2} \\ %
        \rowcolor{baselinecolor}
        \mvit-B & \textbf{84.4} & 253 & \textbf{51.0}  &  2.1 & 7.2  & 6.9 \\
    \end{tabular}
    \vspace{-10pt}
    \caption{\textbf{Runtime comparison on IN-1K and COCO.} %
    We report accuracy and throughput on IN-1K, measured with a V100 GPU as in~\cite{liu2021swin}. COCO models are measured similarly and also for training throughput and memory. Batch size for all measures is identical.  %
    }
    \label{tab:coco_runtime}
    \vspace{-15pt}
\end{table}

\paragraph{Runtime comparison.} We conduct a runtime comparison for \mvit and Swin~\cite{liu2021swin} in Table~\ref{tab:coco_runtime}. We see that \mbox{\mvit-S} surpasses Swin-B on both IN-1K (+\textbf{0.3\%}) and COCO (+\textbf{1.4\%}) while having a higher throughput (341 im/s \vs 276 im/s) on IN-1K and also trains faster (2.7iter/s \vs 2.5iter/s) on COCO with less memory cost (5.2G \vs 6.3G). \mvit-B is slightly slower but significantly more accurate (+1.1\% on IN-1K and +2.5\boxAP on COCO).

\paragraph{Single-scale \vs multi-scale for detection.} Table~\ref{tab:ablation:fpn} compares the default multi-scale (FPN) detector with the single-scale detector for ViT-B and \mvit-S. As ViT produces feature maps at a single scale in the backbone, we adopt a simple scheme~\cite{li2021benchmarking} to up-/downsample features to integrate with FPN. For single-scale, we directly apply the detection heads to the last Transformers block. 
\begin{table}[h]
	\vspace{-5pt}
	\centering
	\small
	\tablestyle{2pt}{1.02}
	\begin{tabular}{ll|ccc}
		variant &  FPN & AP$^\text{box}$ & AP$^\text{mask}$ & FLOPs (G) \\
		\shline
		ViT-B & no                  & 45.1  &	40.6 & 725 \\
		ViT-B & yes                 & \textbf{46.6}  &	\textbf{42.3} & 879 \\
        \hline
		\mvit-S              & no     & 47.0  &	41.4 & 276\\
		\rowcolor{baselinecolor}
		\mvit-S & \default{yes}    & \textbf{49.9}	&   \textbf{45.1} & 326 \\
	\end{tabular}
	\vspace{-10pt}
	\caption{\textbf{Single-scale \vs Multi-scale (FPN) on COCO.} ViT-B and \mvit-S models are equipped with or w/o a feature pyramid network (FPN). Both FPN models outperforms their single-scale variant while while \mvit achieves even larger gains. %
	\label{tab:ablation:fpn}
	} 
	\vspace{-13pt}
\end{table}

As shown in Table~\ref{tab:ablation:fpn}, FPN significantly improves performance for both backbones while \mvit-S is consistently better than ViT-B. Note that the FPN gain for \mvit-S (+2.9 \boxAP ) is much larger than those for ViT-B (+1.5 \boxAP), which shows the effectiveness of a native hierarchical multi-scale design for dense object detection tasks.

\section{Experiments: Video Recognition}
\label{sec:exp:short}

We apply our \mvit on Kinetics-400~\cite{Kay2017} (K400), Kinetics-600 (K600)~\cite{Carreira2017}, and Kinetics-700 (K700)~\cite{Carreira19} and Something-Something-v2~\cite{ssv2} (SSv2) datasets.%

\paragraph{Settings.} By default, our \mvit models are trained \emph{from scratch} on Kinetics and fine-tuned from Kinetics models for SSv2. The training recipe and augmentations follow \cite{MViT,fan2020pyslowfast}. When using IN-1K or IN-21K as pre-training, we adopt the initialization scheme introduced in \sref{sec:mvit_v2_video} and shorter training.

For the temporal domain, we sample a $T\times\tau$ clip from the full-length video which contains $T$~frames with a temporal stride of $\tau$. For inference, we follow testing strategies in \cite{Feichtenhofer2019,MViT} and get final score by averaged from sampled temporal clips and spatial crops. Implementation and training details are in \sref{app:sec:details}.

\subsection{Main Results}

\paragraph{Kinetics-400.} \tblref{tab:sota:k400} compares \mvit to prior work, including state-of-the-art CNNs and ViTs.

When training from scratch, our \mvit-S \& B models produce 81.0\% \& 82.9\% top-1 accuracy which is \textbf{+2.6}\% \& \textbf{+2.7}\% higher than their MViTv1~\cite{MViT} counterparts. These gains stem solely from the improvements in \sref{sec:mvit_v2_improve}, as the training recipe is identical. %

Prior ViT-based models require large-scale pre-training on IN-21K to produce best accuracy on K400. We fine-tune our \mvit-L with large spatiotemporal input size 40\x312$^2$ (\textit{time} \x \textit{space}$^2$) to reach 86.1\% top-1 accuracy, showing the performance of our architecture in a large-scale setting.

\begin{table}[t!]
		\vspace{-15pt}
	\hspace*{-7pt}
	\centering
	\tablestyle{2.0pt}{1.08}
		\resizebox{1.04\linewidth}{!}{
	\begin{tabular}{l|c|c|c|r|r}
		\multicolumn{1}{c|}{model}  &\multicolumn{1}{c|}{pre-train} &   top-1  & top-5  & {\scriptsize FLOPs\x views}  & Param \\
		\shline
		
		\hline		
		{SlowFast} {\scriptsize{16\x 8 +NL}} \cite{Feichtenhofer2019}    &- &{79.8} & {93.9}  & 234\x3\x10 & 59.9  \\
		X3D-XL~\cite{feichtenhofer2020x3d}  &- & 79.1  & {93.9} &48.4\x3\x10 & 11.0  \\ 
		MoViNet-A6~\cite{kondratyuk2021movinets} & - & 81.5 & 95.3 & 386\x1\x1 & 31.4 \\
		\hline
        MViTv1, 16\x4~\cite{MViT} & - & 78.4 & 93.5 & 70.3\x1\x5   & 36.6 \\
	    MViTv1, 32\x3~\cite{MViT} & - & 80.2 & 94.4 & 170\x1\x5   & 36.6 \\

        \rowcolor{baselinecolor}
		\textbf{\mvit}-S, 16\x4  & - & 81.0  & 94.6&  64\x1\x5   & 34.5 \\ %
        \rowcolor{baselinecolor}
		\textbf{\mvit}-B, 32\x3 & - & \textbf{82.9} & \textbf{95.7} & 225\x1\x5   & 51.2 \\ %
        \shline
		{ViT-B-VTN \cite{neimark2021video}}  &  \multirow{6}{*}{\scriptsize{IN-21K}} &    \demph{78.6} &  \demph{93.7} &  {4218\x1\x1} &  {114.0} \\
		{ViT-B-TimeSformer \cite{bertasius2021space}}  &   &  \demph{80.7} &  \demph{94.7} &  2380\x3\x1 &  {121.4} \\ 
		{ViT-L-ViViT \cite{arnab2021vivit}}  &   &  \demph{81.3} &  \demph{94.7} &  3992\x3\x4 & 310.8 \\ 
		Swin-L$\uparrow$ 384$^{2}$~\cite{liu2021video}&   &  \demph{84.9} &  \demph{96.7} &  2107\x5\x10 & 200.0 \\ 
        \rowcolor{baselinecolor}
        \rowcolor{baselinecolor}
		\textbf{\mvit}-L$\uparrow$ 312$^{2}$, 40\x3 &  & \demph{\textbf{86.1}} & \demph{\textbf{97.0}} & 2828\x3\x5   & 217.6\\ 
	
	    \end{tabular}
	}
	\vspace{-10pt}
	\caption{\textbf{Comparison with previous work on Kinetics-400}. We report the inference cost with a single ``view" (temporal clip with spatial crop) $\times$ the number of views (FLOPs\x view$_\text{space}$\x view$_\text{time}$). Magnitudes are Giga ($10^9$) for FLOPs and Mega ($10^6$) for Param. 
	}
	\label{tab:sota:k400}
	\vspace{-10pt}
\end{table}

\begin{table}[t!]
	\centering
	\small
	\tablestyle{2pt}{1.05}
	\begin{tabular}{l|c|c|c|r|r}
		\multicolumn{1}{c|}{model}  &\multicolumn{1}{c|}{pretrain} &   top-1  & top-5  & FLOPs\x views  & Param \\
		\shline
		{SlowFast} {\scriptsize{16\x 8 +NL}} \cite{Feichtenhofer2019}  & - & {81.8}  & {95.1} &  234\x3\x10 & 59.9  \\
		{X3D-XL}~\cite{feichtenhofer2020x3d} &   -  & {81.9}  & {95.5} & 48.4\x3\x10 & 11.0  \\
		MoViNet-A6~\cite{kondratyuk2021movinets} & - & 84.8 & 96.5 & 386\x1\x1 & 31.4 \\
		MViTv1-B-24, 32\x3~\cite{MViT} & - & 84.1 & 96.5 & 236\x1\x5  & 52.9 \\
        \hline
        \rowcolor{baselinecolor}
		\textbf{\mvit}-B, 32\x3 & - & \textbf{85.5} & \textbf{97.2} & 206\x1\x5  & 51.4 \\ %
        \shline
		ViT-L-ViViT \cite{arnab2021vivit}  & \multirow{5}{*}{\scriptsize{IN-21K}} &  \demph{83.0} &  \demph{95.7} &  3992\x3\x4 & 310.8 \\ 
		
		Swin-B~\cite{liu2021video}&   &  \demph{84.0} &  \demph{96.5} &  282\x3\x4 & 88.1 \\ 
		Swin-L$\uparrow$ 384$^{2}$~\cite{liu2021video}&   &  \demph{86.1} &  \demph{97.3} &  2107\x5\x10 & 200.0 \\
        \rowcolor{baselinecolor}
		\textbf{\mvit}-L$\uparrow$ 312$^{2}$, 32\x3 &   & \demph{{87.2}} & \demph{{97.6}}  & 2063\x3\x4   & 217.6\\ %
    \rowcolor{baselinecolor}
			\textbf{\mvit}-L$\uparrow$ 312$^{2}$, 40\x3 &  & 
			\demph{{87.5}} & \demph{{97.8}}  & 2828\x3\x4   & 217.6\\ %
    \rowcolor{baselinecolor}
			\textbf{\mvit}-L$\uparrow$ 352$^{2}$, 40\x3 &  & 
		\demph{\textbf{87.9}} & \demph{\textbf{97.9}} & 3790\x3\x4   & 217.6\\ %
	\end{tabular}
		\vspace{-10pt}
	\caption{\textbf{Comparison with previous work on Kinetics-600}. ~~~~~~~~~~~~~~~~~ %
	}
	\label{tab:sota:k600}
\vspace{-10pt}
\end{table}

\begin{table}[t]
	\vspace{-10pt}
	\centering
	\small
	\tablestyle{2pt}{1.05}
	\begin{tabular}{l|c|c|c|r|r}
		\multicolumn{1}{c|}{model}  &\multicolumn{1}{c|}{pretrain} &   top-1  & top-5  & FLOPs\x views  & Param \\
		\shline
		{SlowFast} {\scriptsize{16\x 8 +NL}}~\cite{Feichtenhofer2019}  & K600 & {71.0}  & 89.6 & 234\x3\x10 & 59.9  \\
		MoViNet-A6~\cite{kondratyuk2021movinets} & N/A & 72.3 & N/A & 386\x1\x1 & 31.4 \\
	    \rowcolor{baselinecolor}
			\textbf{\mvit}-B, 32\x3 & - & 
			\demph{{76.6}} & \demph{{93.2}}  & 206\x3\x3   & 51.4\\ %
    \rowcolor{baselinecolor}
			\textbf{\mvit}-L$\uparrow$ 312$^{2}$, 40\x3 & \scriptsize{IN-21K} & 
			\demph{\textbf{79.4}} & \demph{\textbf{94.9}}  & 2828\x3\x3   & 217.6\\ %
	\end{tabular}
		\vspace{-10pt}
	\caption{\textbf{Comparison with previous work on Kinetics-700}. ~~~~~~~~~~~~~~~~~ %
	}
	\label{tab:sota:k700}
\vspace{-25pt}
\end{table}

\paragraph{Kinetics-600/-700.} \tblref{tab:sota:k600} shows the results on K600. We train \mvit-B, 32\x3 from scratch and achieves 85.5\% top-1 accuracy, which is better than the MViTv1 counterpart ({+1.4\%}), 
and even better than other ViTs with IN-21K pre-training(\eg +1.5\% over Swin-B~\cite{liu2021video}) while having \app2.2\x and \app40\% fewer FLOPs and  parameters. The larger \mvit-L 40\x3 sets a new state-of-the-art at \textbf{87.9}\%. 

In \tblref{tab:sota:k700}, our \mvit-L achieves \textbf{79.4}\% on K700 which greatly surpasses the previous best result by \textbf{+7.1}\%.

\paragraph{Something-something-v2.} \tblref{tab:sota:ssv2} compares methods on a more `temporal modeling' dataset SSv2. Our \mvit-S with 16 frames first improves over MViTv1 counterpart by a large gain ({+3.5\%}), which verifies the effectiveness of our proposed pooling attention for temporal modeling. The deeper \mvit-B achieves 70.5\% top-1 accuracy, surpassing the previous best result Swin-B with IN-21K and K400 pre-training by {+0.9\%} while using ~\app 30\% and 40\% fewer FLOPs and parameters and only K400. With IN-21K pretraining, \mbox{\mvit-B}  boosts accuracy by 1.6\% and achieves \textbf{72.1}\%. \mvit-L achieves \textbf{73.3}\% top-1 accuracy. %
	\begin{table}[t] 
			\vspace{-15pt}
		\centering
		\small
		\tablestyle{2pt}{1.02}
		\resizebox{1\linewidth}{!}{
			\begin{tabular}{l|c|c|c|r|r}
				\multicolumn{1}{c|}{model}  &\multicolumn{1}{c|}{pretrain} &   top-1  & top-5  & FLOPs\x views  & Param \\
				\shline
				TEA~\cite{li2020tea}    & IN-1K & 65.1 & 89.9 & 70\x3\x10 & -\\
				MoViNet-A3~\cite{kondratyuk2021movinets} & N/A & 64.1 & 88.8 & 24\x1\x1 & 5.3 \\
				{ViT-B-TimeSformer~\cite{bertasius2021space}}  &  IN-{21K} &  62.5 &  - &  1703\x3\x1 &  {121.4}  \\
               {MViTv1}-B-24, 32\x3 & { K600} & {68.7}	& {91.5} & 236.0\x3\x1 & 53.2\\ %
                \hline
				SlowFast R101, 8\x8~\cite{Feichtenhofer2019} & \multirow{5}{*}{K400}& 63.1 & 87.6 & 106\x3\x1 & 53.3  \\
				MViTv1-B, 16\x4 & & 64.7	& 89.2 & 70.5\x3\x1 & 36.6 \\ %
				MViTv1-B, 64\x3 & & {67.7}	& {90.9} & 454\x3\x1 & 36.6\\ %
                \rowcolor{baselinecolor}
				\textbf{\mvit}-S, 16\x4 & & 68.2	& 91.4 & 64.5\x3\x1 & 34.4 \\ 
                \rowcolor{baselinecolor}
				\textbf{\mvit}-B, 32\x3 & & \textbf{70.5}	& \textbf{92.7} & 225\x3\x1 & 51.1\\ 
                \hline
                Swin-B~\cite{liu2021video} & {\scriptsize{IN{21K} + K400}} & 69.6 & 92.7 & 321\x3\x1 & 88.8 \\
                \rowcolor{baselinecolor}
                \textbf{\mvit}-B, 32\x3 &  {\scriptsize{IN{21K} + K400}} & {72.1}	& {93.4} & 225\x3\x1 & 51.1\\ 

            \rowcolor{baselinecolor}
            \textbf{\mvit}-L$\uparrow$ 312$^{2}$, 40\x3 & {\scriptsize{IN{21K} + K400}} & {\textbf{73.3}} & {\textbf{94.1}} & 2828\x3\x1   & 213.1\\ 
			\end{tabular}
		}
		\vspace{-10pt}
		\caption{\textbf{Comparison with previous work on SSv2}.~~~~~~~~~~~~~~~~~
		}
\vspace{-15pt}
		\label{tab:sota:ssv2}
	\end{table}

\subsection{Ablations on Kinetics}\label{sec:exp_video_ablation}

In this section, we carry out \mvit ablations on K400. The video ablation our technical improvements share trends with \tblref{tab:ablation:pos} \& \ref{tab:ablation:residual} and are in~\sref{app:sec:k400_ablation}.

	\begin{table}[h]
		\centering
		\small
		\vspace{-10pt}
		\tablestyle{3pt}{1.01}
		\begin{tabular}{l|x{22}|x{22}|x{22}x{22}|x{22}r}
			\multicolumn{1}{c|}{model } & T\x$\tau$ & scratch & IN1k & IN21k & FLOPs & Param \\
			\shline
			\mvit-S  & 16\x4 & 81.2  & 82.2 & 82.6 & 64 & 34.5 \\ %
			\mvit-B  & 32\x3 & 82.9  & 83.3 & 84.3 & 225 & 51.2 \\ %
			\mvit-L  &  40\x3 & 81.4  & 83.4 & 84.5 & 1127 & 217.6\\ %
			\mvit-L$\uparrow$ 312$^{2}$   &  40\x3 &  81.8  & 84.4 & 85.7 & 2828 & 217.6 \\ %
		\end{tabular}
		\vspace{-10pt}
		\caption{\textbf{Effect of pre-training on K400}. We use view$_\text{space}$\x view$_\text{time}$ = 1\x10 crops for inference. } %
		\label{tab:ablation:k400_pertraining} 
		\vspace{-13pt}
	\end{table}

\paragraph{Effect of pre-training datasets.} \tblref{tab:ablation:k400_pertraining} compares the effect different pre-training schemes on K400. We observe that: (\textit{i}) For \mvit-S and \mvit-B models, using either IN1K or IN21k pre-training boosts accuracy compared to \emph{training from scratch},  \eg \mvit-S gets \textbf{+1.0\%} and \textbf{1.4\%} gains with IN1K and IN21K pre-training. (\textit{ii}) For large models, ImageNet pre-training is necessary as they are heavily overfitting when trained from scratch (\cf \tblref{tab:sota:k400}). %

\section{Conclusion}

We present an improved Multiscale Vision Transformer as a general hierarchical architecture for visual recognition. In empirical evaluation, MViT shows strong performance compared to other vision transformers and achieves state-of-the-art accuracy on widely-used benchmarks across image classification, object detection, instance segmentation and video recognition. We hope that our architecture will be useful for further research in visual recognition.

\clearpage
\appendix

\setcounter{table}{0}
\renewcommand{\thetable}{A.\arabic{table}}
\renewcommand{\thefigure}{A.\arabic{figure}}

\section*{Appendix}\label{sec:details}
This appendix provides further details for the main paper:

\sref{sec:resultsapp} contains further \textit{results} for COCO object detection (\sref{app:sec:coco_results}) AVA action detection (\sref{app:sec:ava_results}) and  ImageNet classification (\sref{app:sec:in_results}), as well as \emph{ablations} for ImageNet classification and COCO object detection (\sref{app:sec:in_coco_ablation}) and Kinetics action classification (\sref{app:sec:k400_ablation}).

 \sref{app:sec:details} contains additional \mvit \emph{upgrade details} (\sref{app:sec:upgrades}), and additional \textit{implementation details} for: ImageNet classification (\sref{app:sec:in_details}), COCO object detection (\sref{app:sec:coco_details}), Kinetics action classification (\sref{app:sec:kinetics_details}), SSv2 action classification (\sref{app:sec:ssv2_details}), and AVA action detection (\sref{app:sec:ava_details}).

\section{Additional Results} \label{sec:resultsapp}

\subsection{Results: COCO Object Detection} \label{app:sec:coco_results}

\paragraph{System-level comparsion on COCO.} Table~\ref{app:tab:coco_system_results} shows the system-level comparisons on COCO data. We compare our results with previous state-of-the-art models. We adopt SoftNMS~\cite{bodla2017soft} during inference, following~\cite{liu2021swin}. \mvit-L$^*$ achieves 58.7 \boxAP with multi-scale testing, which is already +0.7 AP better than the best results of Swin-L$^*$ that relies on the improved HTC++ detector~\cite{liu2021swin}.

\begin{table}[h!] 
    \centering
    \tablestyle{3pt}{1.1}
    \begin{tabular}{lc|cc|cr}
        model & framework & {AP$^\text{box}$} & {AP$^\text{mask}$} & Flops & Param \\
            \shline
            Copy-Paste~\cite{ghiasi2021simple} & \scriptsize{Cascade, NAS-FPN} & 55.9 & 47.2 & 1440 & 185 \\
            Swin-L\cite{liu2021swin} & HTC++ & 57.1 & 49.5 & 1470 & 284 \\
            Swin-L~\cite{liu2021swin}$^*$ & HTC++ & 58.0 & 50.4 & - & 284 \\
            \hline
            \rowcolor{baselinecolor}
            \textbf{\mvit}-L  & Cascade   & 56.9	 & 48.6	& 1519 & 270 \\
            \rowcolor{baselinecolor}
            \textbf{\mvit}-L$^*$ &  Cascade     & \textbf{58.7}	&\textbf{50.5} & - & 270 \\
        \end{tabular}
        \vspace{-5pt}
        \caption{\textbf{System-level comparison on COCO object detection and segmentation}. The detection frameworks include Cascade Mask R-CNN~\cite{cai2018cascade} (Cascade), the improved Hybrid Task Cascade (HTC++)~\cite{liu2021swin} and Cascade Mask R-CNN with NAS-FPN~\cite{ghiasi2019fpn}. $^*$~indicates multi-scale testing.
        FLOPs and Params are in Giga ($10^9$) and Mega ($10^6$).
        }
        \vspace{-15pt}
        \label{app:tab:coco_system_results}
\end{table}

\subsection{Results: AVA Action Detection} \label{app:sec:ava_results}

\paragraph{Results on AVA.} \tblref{tab:sota:ava} shows the results of our \mvit models compared with prior state-of-the-art works on the AVA dataset~\cite{Gu2018} which is a dataset for spatiotemporal-localization of human actions. 

We observe that MViT consistently achieves better results compared to MViTv1~\cite{MViT} counterparts. For example, \mvit-S 16\x4 ($26.8$ mAP) improves +$2.3$ over MViTv1-B 16\x4 ($24.5$ mAP) with fewer flops and parameters (both with the same recipe and default K400 pre-training). 
For K600 pre-training, \mvit-B 32\x3 ($29.9$ mAP) improves +$1.2$ over MViTv1-B-24 32\x3. This again validates the effectiveness of the proposed \mvit improvements in \S{\color{red}4.1} of the main paper. Using \emph{full-resolution} testing (without cropping) can further improve \mvit-B by +$0.6$ to achieve $30.5$ mAP. Finally, the larger \mvit-L 40\x3 achieves the state-of-the-art results at $34.4$ mAP using IN-21K and K700 pre-training.

\subsection{Results: ImageNet Classification} \label{app:sec:in_results}

\paragraph{Results of ImageNet-1K.} \tblref{tab:sota:in1k_full} shows the comparison of our \mvit with \emph{more} prior work (without external data or distillation models) on ImageNet-1K. As shown in the Table, our \mvit achieves better results than any previously published methods for a variety of model complexities. We note that our improvements to pooling attention bring significant gains over the MViTv1~\cite{MViT} counterparts which use exactly the same training recipes (for all datasets we compare on); therefore the gains over MViTv1 stem solely from our technical improvements in \S{\color{red}4.1} of the main paper.

\begin{table}[t!]
		\vspace{-15pt}
	\centering
	\tablestyle{2.5pt}{1.05}
	\begin{tabular}{l|c|cc|c|r}
        \multirow{2}{*}{{}} & \multirow{2}{*}{{}} &   \multicolumn{2}{c|}{val mAP}  &  \multirow{2}{*}{} &  \multirow{2}{*}{} \\ 
        model&  pretrain &   center  & full & FLOPs & Param\\
		\shline
		{SlowFast}, 4\x16, R50 \cite{Feichtenhofer2019} & \multirow{6}{*}{K400} & 21.9 & - & 52.6 & 33.7 \\ %
		SlowFast, 8\x8, R101~\cite{Feichtenhofer2019}   &  & 23.8 & - & 137.7 & 53.0 \\  %
		{MViTv1}-B, 16\x4~\cite{MViT}   &  & 24.5 & - & 70.5 & 36.4 \\ %
		{MViTv1}-B, 64\x3~\cite{MViT}    &   & {27.3} & - &454.7 & 36.4 \\  %
        \rowcolor{baselinecolor}
        \textbf{\mvit}-S, 16\x4            &    &  26.8   & 27.6 & 64.5   & 34.3 \\ %
        \rowcolor{baselinecolor}
        \textbf{\mvit}-B, 32\x3            &    &  \textbf{28.1}   & \textbf{29.0} & 225.2   & 51.0 \\ %
		\hline
		
		\scriptsize{{SlowFast}, 8\x8 R101+NL}~\cite{Feichtenhofer2019}  & \multirow{8}{*}{K600}   &  27.1 & - & 146.6 & 59.2 \\ %
		\scriptsize{{SlowFast}, 16\x8 R101+NL}~\cite{Feichtenhofer2019}   &    &  27.5 & - & 296.3 & 59.2 \\
		X3D-XL~\cite{feichtenhofer2020x3d}  &    &  27.4 & - & 48.4 & 11.0 \\
        Object Transformer~\cite{wu2021towards}  & & \textbf{31.0} & - & 243.8 & 86.2 \\
        ACAR 8\x8, R101-NL~\cite{pan2021actor}              & & - & \textbf{31.4} & N/A & N/A \\
		{MViTv1}-B, 16\x4~\cite{MViT}   &  & 26.1 & - & 70.4 & 36.3 \\ %
		{MViTv1}-B-24, 32\x3~\cite{MViT}    &   & {28.7} && 236.0 & 52.9 \\  %
        \rowcolor{baselinecolor}
        \textbf{\mvit}-B, 32\x3                 & & 29.9   & 30.5 &  225.2   & 51.0\\
        \hline
        ACAR 8\x8, R101-NL~\cite{pan2021actor}              & \multirow{1}{*}{K700} & - & \textbf{33.3} & N/A & N/A \\
        \rowcolor{baselinecolor}
        \textbf{\mvit}-B, 32\x3                 & K700 & \textbf{31.3} & 32.3 & 225.2   & 51.0 \\
        \hline
        \rowcolor{baselinecolor}
        \textbf{\mvit}-L$\uparrow$ 312$^{2}$, 40\x3  & \scriptsize{IN21K+K700} & \textbf{33.5} & \textbf{34.4} & 2828 & 213.0 \\
		
	\end{tabular}
	\vspace{-1.0em}
	
	\caption{\textbf{Comparison with previvous work on AVA v2.2}. We adopt two test strategies: 1) \emph{center} (\emph{single center crop}): we resize the shorter spatial side to 224 pixels and takes a 224$^2$ center crop for inference. 2) \emph{full} (\emph{full-resolution}): we resize the shorter spatial side to 224 pixels and take the full image for inference. We report inference cost with the \emph{center} testing strategy (\ie 224$^2$ input). Magnitudes are Giga ($10^9$) for FLOPs and Mega ($10^6$) for Param. 
	}
	\label{tab:sota:ava}
\end{table}

\subsection{Ablations: ImageNet and COCO} \label{app:sec:in_coco_ablation}

\begin{table}[t!]
    \vspace{-32pt}
    \centering
    \tablestyle{2.4pt}{1.0}
    \begin{tabular}{l|x{25}x{25}|r|r}
         &   \multicolumn{2}{c|}{Acc}   &   &  \\
        \multicolumn{1}{c|}{model}  & center & resize  & FLOPs (G)   & Param (M) \\
        \shline
    
        {RegNetY-4GF \cite{ilija_2020} }        &  80.0 && 4.0 &  21 \\ 
        {RegNetZ-4GF \cite{dollar2021fast} }    &  \textbf{83.1} && 4.0 &  28 \\ 
        {EfficientNet-B4 $\uparrow$ 380$^2$\cite{tan2019efficientnet} }  &   82.9 && 4.2 & 19 \\
        \hline
        DeiT-S~\cite{touvron2020training}           &  79.8 && 4.6 & 22 \\
        PVT-S~\cite{wang2021pyramid}                &  79.8 && 3.8 & 25 \\
        TNT-S~\cite{han2021tnt}                     & 81.5 && 5.2 & 24 \\
        T2T-ViT$_t$-14~\cite{yuan2021tokens}       &  81.7 && 6.1 & 22 \\
        CvT-13~\cite{wu2021cvt}                     & 81.6 && 4.5 & 20 \\
        Twins-S~\cite{chu2021twins}                 & 81.7 && 2.9 & 24 \\
        ViL-S-RPB~\cite{zhang2021multi}                 & 82.4 && 4.9 & 25 \\
        PVTv2-V2~\cite{wang2021pvtv2}              & 82.0 && 4.0 & 25    \\
        CrossViT$_{c}$-15~\cite{chen2021crossvit}  &  82.3  && 6.1 & 28 \\
        XCiT-S12~\cite{el2021xcit}                  & 82.0 && 4.8 & 26 \\
        Swin-T~\cite{liu2021swin}                   & 81.3 && 4.5 & 29  \\
        CSWin-T~\cite{dong2021cswin}                &  82.7 && 4.3  & 23 \\
        \rowcolor{baselinecolor}
        \textbf{\mvit}-T                          & 82.3 && 4.7 & 24 \\   %
        \shline
        {RegNetY-8GF \cite{ilija_2020} }  &   81.7 && 8.0 &  39 \\
        {EfficientNet-B5 $\uparrow$ 456$^2$\cite{tan2019efficientnet}}  &   \textbf{83.6} && 9.9 & 30\\
        \hline
        
        PVT-M~\cite{wang2021pyramid}               & 81.2 && 6.7 & 44    \\
        T2T-ViT$_t$-19~\cite{yuan2021tokens}       &  82.4 && 9.8 & 39 \\
        CvT-21~\cite{wu2021cvt}                    & 82.5 && 7.1 & 32    \\
        Twins-B~\cite{chu2021twins}                & 83.2 && 8.6 & 56    \\
        ViL-M-RPB~\cite{zhang2021multi}               & 83.5 && 8.7 & 40 \\
        PVTv2-V2-B3~\cite{wang2021pvtv2}              & 83.2 && 6.9 & 45    \\
        CrossViT$_{c}$-18~\cite{chen2021crossvit}  &  82.8  && 9.5 & 44 \\
         XCiT-S24~\cite{el2021xcit}                  & 82.6 && 9.1 & 48\\
        Swin-S~\cite{liu2021swin}                  & 83.0 && 8.7 & 50 \\
        CSWin-S~\cite{dong2021cswin}            & \textbf{83.6} && 6.9 & 35 \\
        MViT-v1-B-16~\cite{MViT} &     83.0 && 7.8 & 37 \\
        \rowcolor{baselinecolor}
        \textbf{\mvit}-S &  \textbf{83.6} && 7.0 & 35 \\   %
        
        \shline
        {RegNetY-16GF \cite{ilija_2020} }       &    82.9 && 15.9 &  84 \\
        {RegNetZ-16GF \cite{dollar2021fast} }   &    84.1 && 15.9 &  95 \\
        {EfficientNet-B6 $\uparrow$ 528$^2$\cite{tan2019efficientnet}}  &    84.2 && 19 & 43 \\
        NFNet-F0 $\uparrow$ 256$^2$\cite{brock2021high}                         &      83.6 && 12.4 & 72 \\ 
        \hline
        DeiT-B \cite{touvron2020training}           &      81.8 && 17.6 & 87 \\
        PVT-L~\cite{wang2021pyramid}               & 81.7 && 9.8 & 61    \\
        T2T-ViT$_t$-21~\cite{yuan2021tokens}       &  82.6 && 15.0 & 64 \\
        TNT-B~\cite{han2021tnt}                    & 82.9 && 14.1  &  66\\
        Twins-L~\cite{chu2021twins}                & 83.7 && 15.1  &  99\\
        ViL-B-RPB~\cite{zhang2021multi}                 & 83.7 && 13.4  & 56 \\
        PVTv2-V2-B5~\cite{wang2021pvtv2}              & 83.8 && 11.8 & 82    \\
        CaiT-S36~\cite{touvron2021going}               & 83.3 && 13.9 & 68 \\
        XCiT-M24~\cite{el2021xcit}                  & 82.7 && 16.2 & 84\\
        Swin-B~\cite{liu2021swin}              & 83.3   && 15.4  & 88   \\
        CSWin-B~\cite{dong2021cswin}                                 & 84.2 && 15.0    & 78  \\
        {MViTv1}-B-24~\cite{MViT} &     83.4 && 10.9 & 54 \\
        \rowcolor{baselinecolor}
        \textbf{\mvit}-B &  \textbf{84.4} && 10.2 & 52 \\  %
        \shline
        {EfficientNet-B7 $\uparrow$ 600$^2$\cite{tan2019efficientnet}}  &  84.3 && 37.0 & 66 \\
        NFNet-F1 $\uparrow$ 320$^2$\cite{brock2021high}                         &      84.7 && 35.5 & 133 \\ 
        \hline 
        DeiT-B $\uparrow$ 384$^{2}$ \cite{touvron2020training} &  83.1 &&  55.5 &   87\\
        TNT-B $\uparrow$ 384$^{2}$ ~\cite{han2021tnt}                    & 83.9 && N/A  &  66\\
        CvT-32 $\uparrow$ 384$^{2}$~\cite{wu2021cvt}                    && 83.3 & 24.9 & 32    \\
        CaiT-S36$\uparrow$ 384$^{2}$~\cite{touvron2021going}   && 85.0 & 48 & 68 \\
        Swin-B $\uparrow$ 384$^{2}$ \cite{liu2021swin} && 84.2 &  47.0 &   88 \\
        MViT-v1-B-24 $\uparrow$ 320$^{2}$~\cite{MViT} &  {84.8}  &&  32.7 & 73 \\ %
        \rowcolor{baselinecolor}
        \textbf{\mvit}-B $\uparrow$ 384$^{2}$ &  \textbf{85.2} & \textbf{85.6}&	36.7 & 52  \\  %
        \shline
        NFNet-F2 $\uparrow$ 352$^2$\cite{brock2021high}                         &      85.1 && 62.6 & 194 \\ 
        \hline
        XCiT-M24~\cite{el2021xcit}                  & 82.9 && 36.1 & 189\\
        CoAtNet-3~\cite{dai2021coatnet}             & 84.5 && 34.7 & 168 \\
        \rowcolor{baselinecolor}
        \textbf{\mvit}-L  &    \textbf{85.3}  &&  42.1 & 218  \\
        \shline
        NFNet-F4 $\uparrow$ 512$^2$\cite{brock2021high}                         &      85.9 && 215.3 & 316 \\ 
        \hline
        CoAtNet-3~\cite{dai2021coatnet} $\uparrow$ 384$^{2}$  & & 85.8 & 107.4 & 168 \\
        \rowcolor{baselinecolor}
        \textbf{\mvit}-L $\uparrow$ 384$^{2}$  &    \textbf{86.0}  & \textbf{86.3}&  140.2 & 218 \\
    \end{tabular}%
    \vspace{-10pt}
    \caption{\textbf{Comparison to previous work on ImageNet-1K}. Input images are 224$\times$224 by default and $\uparrow$ denotes using different sizes. MViT is trained for 300 epochs without any external data or models. We report our $\uparrow$ 384$^{2}$ models tested using a \textit{center} crop or a \textit{resized} \textit{full}  crop of the original image, to compare to prior work.
    }
    \label{tab:sota:in1k_full}
    \vspace{-30pt}
\end{table}

\paragraph{Decomposed relative position embeddings.} As introduced in Sec.~\ref{sec:mvit_v2_improve}, our Relative position embedding is only applied for $Q_i$ by default. We could further extend it to all Q, K and V terms for attention layers:
\begin{align*}
	& \mathrm{Attn}(Q, K, V) =  AV + E^{(\mathrm{rel_v})}, \\[0pt]
\text{where \quad} A & = \mathrm{Softmax}\rbr{(QK^\top + E^{(\mathrm{rel_q})} + E^{(\mathrm{rel_k})}) / \sqrt{d}}.
\end{align*}
And the rel pos terms are defined as:
\begin{align*}
E^{(\mathrm{rel_q})}_{ij} = & Q_i \cdot R^{q}_{p(i),p(j)}, \\[0pt]
E^{(\mathrm{rel_k})}_{ij} = & R^{k}_{p(i),p(j)} \cdot K_i, \\[0pt]
E^{(\mathrm{rel_v})}_{i} = & \sum_{j} A_{ij} * R^{v}_{p(i),p(j)}. 
\end{align*}
Table~\ref{tab:app:relpos} shows the ablation experiments: different variants achieve similar accuracy  on ImageNet and COCO. However $\mathrm{rel_v}$ requires more GPU memory (\eg 30.8G vs 6.2G on ImageNet and out-of-memory (OOM) on COCO) and has a \app2.9\x lower test throughput on ImageNet. For simplicity and efficiency, we use only $\mathrm{rel_q}$ by default. 

\begin{table}[t]
    \centering
    \tablestyle{3pt}{1.02}

	\begin{tabular}{ccc|ccc|cc}
        \multicolumn{3}{c|}{rel pos} & \multicolumn{3}{c|}{IN-1K} & \multicolumn{2}{c}{COCO} \\
        $\mathrm{rel_q}$ & $\mathrm{rel_k}$ & $\mathrm{rel_v}$ & \multirow{1}{*}{Acc} & Mem(G) & Test (im/s) & \multirow{1}{*}{AP$^\text{box}$} & \multirow{1}{*}{AP$^\text{mask}$} \\
		\hline
		\rowcolor{baselinecolor}
		\cmark & \xmark & \xmark & 83.6 & 6.2 & 316 & 49.9 & 45.0\\
		\xmark & \cmark & \xmark & 83.4 & 6.2 & 321 & 49.7 & 44.8\\
		\cmark & \cmark & \xmark & 83.6 & 6.4 & 300 & 50.0 & 45.0\\
		\xmark & \xmark & \cmark & 83.6	& 30.8 & 109 & OOM  & OOM \\
		\cmark & \xmark & \cmark & 83.7	& 30.9  & 104 & OOM  & OOM \\
		\cmark & \cmark & \cmark & 83.6	& 30.9	& 103 & OOM  & OOM \\
		\hline				
	\end{tabular}
    \vspace{-2pt}
    \caption{\textbf{Ablation of rel pos embeddings} on ImageNet-1K and COCO with MViT-S.
    }
    \label{tab:app:relpos}
    \vspace{-3pt}
\end{table}

\paragraph{Effect of pre-training datasets for detection.} In \S{\color{red}6.2} of the main paper we observe that ImageNet pre-training can have very different effects for different model sizes for video classification. Here, we are interested in the impact of pre-training on the larger IN-21K \vs IN-1K for COCO \textit{object detection tasks}. \tblref{tab:ablation:coco_pretrain} shows our ablation: The large-scale IN-21K pre-training is more helpful for larger models, \eg MViT-B and MViT-L have +0.5 and +0.9 gains in \boxAP. %

\begin{table}[h!]
        \centering
        \tablestyle{2pt}{1.1}
        \begin{tabular}{c|ll|ll}
            \multirow{2}{*}{variant} &  \multicolumn{2}{c|}{AP$^\text{box}$} & \multicolumn{2}{c}{AP$^\text{mask}$}  \\
             \cline{2-5}
               & IN-1k & IN-21k & IN-1k & IN-21k \\
            \shline
            \mvit-S &    49.9 &	 50.2	&45.1 &45.1 \\
            \mvit-B &    51.0 &	 51.5	&45.7 &46.4 \\
            \mvit-L &    51.8 &	 52.7	&46.2 &46.8\\
        \end{tabular}
        \vspace{-2pt}
        \caption{\textbf{Effect of pre-training datasets for COCO.} Detection methods are initialized from IN-1K or IN-21K pre-trained weights. %
        \label{tab:ablation:coco_pretrain}
        } 
        \vspace{-5pt}
    \end{table}

\subsection{Ablations: Kinetics Action Classification} \label{app:sec:k400_ablation}

In \S{\color{red}5.3} of the main paper we ablated the impact of our improvements to pooling attention, \ie  decomposed relative positional embeddings \& residual pooling connections, for image classification and object detection. Here, we ablate the effect of our improvements for video classification. %

\paragraph{Positional embeddings for video.}  \tblref{tab:ablation:k400_pos} compares different positional embeddings for \mvit on K400. Similar to image classification and object detection (Table~{\color{red}6} of the main paper), relative positional embeddings surpass absolute positional embeddings by \app0.6\% comparing (\emph{2}) and (\emph{5}, \emph{6}). Comparing (\emph{5}) to (\emph{6}), our \emph{decomposed space/time rel.}~positional embeddings achieve nearly the same accuracy as the \emph{joint space rel.}~embeddings while being \app\textbf{2}\x~\textit{faster} in training. For \emph{joint space/time rel.} (\emph{5} \vs \emph{7}), our \emph{decomposed space/time rel.} is even \app\textbf{8}\x faster with \app\textbf{2}\x fewer parameters. This demonstrates the effectiveness of our decomposed design for relative positional embeddings.

\begin{table}[t]
    \centering
    \tablestyle{1.8pt}{1.05}
    \begin{tabular}{l|cc|c|ccc}
          & \multicolumn{2}{c|}{rel.~pos.} &  \multirow{2}{*}{abs.~pos.} &  \multirow{1}{*}{Top-1} &  \multirow{1}{*}{Train} & \multirow{1}{*}{Param} \\
        & space  & time          & &  (\%)  & (clip/s)  & (M) \\
        \shline
        (\emph{1}) no pos.&               & &                & 80.1 & \textbf{91.5} & \textbf{34.4}\\
        (\emph{2}) abs.~pos.&             &  & \checkmark    & 80.4 & 91.0 & 34.7\\
        \hline
        (\emph{3}) time-only rel. &     & \checkmark    &    & 80.8 & 80.5 & \textbf{34.4}\\
        (\emph{4}) space-only rel. &{dec.}  &     &          & 80.6 & 76.2 & 34.5\\
        \rowcolor{baselinecolor}
        (\emph{5}) \default{dec. space rel.~+ time rel.} &\default{dec.}        & \default{\checkmark}     & \default{} &81.0  & 66.6 & 34.5 \\
        (\emph{6}) joint space rel.~+ time rel. &joint& \checkmark    & & \textbf{81.1} & 33.6 & 37.1\\
        \demphs{(\emph{7}) joint space/time rel.} & \multicolumn{2}{c|}{\demphs{joint}}   & & \demphs{-} & \demphs{8.4} & \demphs{73.7}\\
    \end{tabular}
    \vspace{-3pt}
    \caption{\textbf{Ablation of positional embeddings} on K400 with \mvit-S 16\x4. Training throughput is measured by average clips per-second with 8 V100 GPUs. Our (\emph{5}) \emph{decomposed space/time rel.}~positional embeddings are accurate and significantly faster than other joint versions. Note that we do not finish the full training for (\emph{7}) \emph{joint space/time rel.}~as the training speed is too slow ($\app$8\x~slower than ours) and (\emph{6}) \emph{joint space rel.}~already shows large drawbacks ($\app$2\x~slower) of joint rel.~positional embeddings.
    }
    \label{tab:ablation:k400_pos}
    \vspace{-1.0em}
\end{table}

\paragraph{Residual pooling connection for video.} \tblref{tab:ablation:k400_res} studies the effect of residual pooling connections on K400. We observe similar results as for image classification and object detection (Table~{\color{red}7} of the main paper), that: both Q pooling blocks and residual paths are \textit{essential} in our improved \mvit and combining them together leads to \textbf{+1.7}\% accuracy on K400 while using them separately only improves slightly (+0.4\%).

\begin{table}[h!]
    \centering
    \tablestyle{1.8pt}{1.05}
    \begin{tabular}{l|cc}
        \multirow{1}{*}{residual pooling} & Top-1 & FLOPs\\
        \shline
        (\emph{1}) w/o                                             & 79.3 & 64 \\
        (\emph{2}) full Q pooling                                  & 79.7 & 65\\
        (\emph{3}) residual                                        & 79.7 & 64\\
        \rowcolor{baselinecolor}
        (\emph{4}) \default{full Q pooling + residual}             & \textbf{81.0} & 65\\
    \end{tabular}
    \caption{\textbf{Ablation of residual pooling connections} on K400 with \mvit-S 16\x4 architecture. %
    }
    \label{tab:ablation:k400_res}
    \vspace{-1.0em}
\end{table}

\section{Additional Implementation Details} \label{app:sec:details}

\subsection{Other Upgrades in MViT} \label{app:sec:upgrades}

Besides the technical improvements introduced in \S{\color{red}4.1} of the main paper, MViT entails two further changes: (i) We conduct the channel dimension expansion in the \textit{attention computation} of the first transformer block of each stage, instead of performing it in the last MLP block of the prior stage as in MViTv1~\cite{MViT}. This change has similar accuracy ($\pm0.1$\%) to the original version, while reducing parameters and FLOPs. (ii) We remove the class token in MViT by default as this has no advantage for image classification tasks. Instead, we average the output tokens from the last transformer block and apply the final classification head upon it. In practice, we find this modification could reduce the training time by \app{8\%}. %

\subsection{Details: ImageNet Classification} \label{app:sec:in_details}

\paragraph{IN-1K training.} We follow the training recipe of MViTv1~\cite{MViT,touvron2020training} for IN-1K training. %
We train for $300$ epochs with 64 GPUs. 
The batch size is 32 per GPU by default. We use truncated normal distribution initialization~\cite{hanin2018start} and adopt synchronized AdamW~\cite{loshchilov2018fixing} optimization with a base learning rate of $\expnum{2}{-3}$ for batch size of 2048. We use a linear warm-up strategy in the first 70 epochs and a decayed half-period cosine schedule~\cite{touvron2020training}. 

For regularization, we set weight decay to $0.05$ for \mvit-T/S/B and $0.1$ for \mvit-L/H and label-smoothing~\cite{Szegedy2015a} to $0.1$. Stochastic depth~\cite{huang2016deep} (\ie drop-path or drop-connect) is also used with rate $0.1$ for \mvit-T \& \mvit-S, rate $0.3$ for \mbox{\mvit-B}, rate $0.5$ for \mvit-L and rate $0.8$ for \mvit-H. Other data augmentations have the same (default) hyperparameters as in \cite{MViT,deit}, including mixup~\cite{zhang2018mixup}, cutmix~\cite{yun2019cutmix}, random erasing~\cite{zhong2020random} and rand augment~\cite{cubuk2020randaugment}.

For 384\x384 input resolution, we fine-tune the models trained on 224\x224 resolution. We decrease the batch size to 8 per GPU and fine-tune 30 epochs with a base learning rate of $\expnum{4}{-5}$ per $256$ batch-size samples.  For \mvit-L and \mvit-H, we disable mixup and fine-tune with a learning rate of $\expnum{5}{-4}$ per batch of $64$.
We linearly scale learning rates with the number of overall GPUs (\ie the overall batch-size).

\paragraph{IN-21K pre-training and fine-tuning on IN-1K.} We download the latest winter-2021 version of IN-21K from the official website. %
The training recipe follows the IN-1K training introduced above except for some differences described next. We train the IN-21K models on the joint set of IN-21K and 1K for 90 epochs (60 epochs for \mvit-H) with a $6.75\times 10^{-5}$ base learning rate for \mvit-S and \mvit-B, and $10^{-4}$ for \mvit-L and \mvit-H, per batch-size of 256. The weight decay is set as 0.01 for \mvit-S and \mvit-B, and 0.1 for \mvit-L and \mvit-H. %

When fine-tuning IN-21K \mvit models on IN-1K for \mvit-L and \mvit-H,
we disable mixup and fine-tune for 30 epochs with a learning rate of $\expnum{7}{-5}$ per batch of $64$. We use a weight decay of $\expnum{5}{-2}$. The \mvit-H $\uparrow$ 512$^2$ model is initialized from the 384$^2$ variant and trained for 3 epochs with mixup enabled and weight decay of $10^{-8}$.

\subsection{Details: COCO Object Detection} \label{app:sec:coco_details} 

For object detection experiments, we adopt two typical object detection framework: Mask R-CNN~\cite{He2017} and Cascade Mask R-CNN~\cite{cai2018cascade} in Detectron2~\cite{wu2019detectron2}. We follow the same training settings from~\cite{liu2021swin}: multi-scale training (scale the shorter side in $[480, 800]$ while longer side is smaller than 1333), AdamW optimizer~\cite{loshchilov2018fixing} ($\beta_1, \beta_2 = 0.9, 0.999$, base learning rate $\expnum{1.6}{-4}$ for base size of 64, and weight decay of 0.1), and 3\x schedule (36 epochs). The drop path rate is set as $0.1, 0.3, 0.4$, $0.5$ and $0.6$ for \mvit-T, \mvit-S, \mvit-B, \mvit-L and \mvit-H, respectively. We use PyTorch's automatic mixed precision during training.

 For the stronger recipe for \mvit-L and \mvit-H in Table.~{\color{red}5} of the main paper, we use \emph{large-scale jittering} (1024\x1024 resolution) as the training augmentation~\cite{ghiasi2021simple} and a longer schedule (50 epochs) with IN-21K pre-training.

\subsection{Details: Kinetics Action Classification} \label{app:sec:kinetics_details}

\paragraph{Training from scratch.} We follow the training recipe and augmentations from \cite{MViT,fan2020pyslowfast} when training from scratch for Kinetics datasets. We adopt synchronized AdamW~\cite{loshchilov2018fixing} and train for 200 epochs with 2 repeated augmentation~\cite{hoffer2020augment} on 128 GPUs. The mini-batch size is 4 clips per GPU. We adopt a half-period cosine schedule \cite{Loshchilov2016} of learning rate decaying. The base learning rate is set as $\expnum{1.6}{-3}$ for 512 batch-size. We use weight decay of $0.05$ and set drop path rate as $0.2$ and $0.3$ for \mvit-S and \mvit-B.

For the input clip, we randomly sample a clip ($T$~frames with a temporal stride of $\tau$; denoted as $T\times\tau$~\cite{Feichtenhofer2019}) from the full-length video during training. For the spatial domain, we use Inception-style~\cite{Szegedy2015} cropping (randomly resize the input \textit{area} between a $[$min, max$]$, scale of $[$0.08,  1.00$]$, and jitter aspect ratio between 3/4 to 4/3). Then we take  an $H\times W$ = 224\x224 crop as the network input.

During inference, we apply two testing strategies following~\cite{Feichtenhofer2019,MViT}: \textit{(i)} Temporally, uniformly samples $K$ clips  (\eg $K$=$5$)  from a video. \textit{(ii)} in spatial axis, scales the shorter spatial side to 256 pixels and takes a 224×224 center crop or 3 crops of 224\x224 to cover the longer spatial axis. The final score is averaged over %
all predictions.

For the input clips, we perform the same data augmentations across all frames, including random horizontal flip, mixup~\cite{zhang2018mixup} and cutmix~\cite{yun2019cutmix}, random erasing~\cite{zhong2020random}, and rand augment~\cite{cubuk2020randaugment}.

For {Kinetics-600} and Kinetics-700, all hyper-parameters are \textit{identical} to K400. %

\paragraph{Fine-tuning from ImageNet.} When using IN-1K or IN-21K as pre-training, we adopt the initialization scheme introduced in \S{\color{red}4.3} of the main paper and shorter training schedules. For example, we train 100 epochs with base learning rate as $\expnum{4.8}{-4}$ for 512 batch-size when fine-tuning from IN-1K for \mvit-S and \mvit-B, and 75 epochs with base learning as $\expnum{1.6}{-4}$ when fine-tuning from IN-21K. For long-term models with 40\x3 sampling, we initialize from the 16\x4 counterparts, disable mixup, train for 30 epochs with learning rate of $\expnum{1.6}{-5}$ at batch-size of 128, and use a weight decay of 10$^{-8}$.

\subsection{Details: Something-Something V2 (SSv2)} \label{app:sec:ssv2_details}

The SSv2 dataset~\cite{ssv2} contains 169k training, and 25k validation videos with 174 human-object interaction classes. We fine-tune the pre-trained Kinetics models and take the same recipe as in \cite{MViT}. Specifically, we train for 100 epochs (40 epochs for \mvit-L) using 64 or 128 GPUs with 8 clips per GPU and a base learning rate of 0.02 (for batch size of 512) with half-period cosine decay \cite{Loshchilov2016}. We adopt synchronized SGD and use weight decay of 10$^{-4}$ and drop path rate of 0.4. The training augmentation is the same as Kinetics in \sref{app:sec:kinetics_details}, except we disable random flipping and repeated augmentations in training.

We use the segment-based input frame sampling~\cite{lin2018temporal, MViT} (split each video into segments, and sample one frame from each segment to form a clip). During inference, we take a single clip with 3 spatial crops to form predictions over a single video.

\subsection{Details: AVA Action Detection} \label{app:sec:ava_details}

The AVA action detection dataset~\cite{Gu2018} assesses the spatiotemporal localization of human actions in videos. It has 211k training and 57k validation video segments. We evaluate methods on AVA v2.2 and use mean Average Precision (mAP) metric on 60 classes as is standard in prior work~\cite{Feichtenhofer2019}.

We use \mvit as the backbone and follow the same detection architecture in \cite{Feichtenhofer2019, MViT} that adapts Faster R-CNN~\cite{Ren2015} for video action detection. Specifically, we extract region-of-interest (RoI) features~\cite{Girshick2015} by frame-wise RoIAlign~\cite{He2017} on the spatiotemporal feature maps from the last \mvit layer. The RoI features are then max-pooled and fed to a per-class, sigmoid classifier for action prediction. 

The training recipe is identical to \cite{MViT} and summarized next. We pre-train our \mvit models on Kinetics. The region proposals are identical to the ones used in \cite{Feichtenhofer2019,MViT}. We use proposals that have overlaps with ground-truth boxes by \mbox{IoU $>$ 0.9} for training. The models are trained with synchronized SGD training on 64 GPUs (8 clips per GPU). The base learning rate is set as $0.6$ with a half-period cosine schedule of learning rate decaying. We train for 30 epochs with linear warm-up~\cite{Goyal2017} for the first 5 epochs and use a weight decay of $\expnum{1}{-8}$ and drop-path rate of $0.4$.

\section{Additional Discussions} \label{app:sec:discussions}

\paragraph{Societal impact.}
Our \mvit is a general vision backbone for various vision tasks, including image recognition, object detection, instance segmentation, video classification and video detection. Though we are not providing any direct applications, it could potentially apply to a wide range of vision-related applications, which then might have a wide range of societal impacts. On the positive side, the better vision backbone could potentially improve the performance of many different computer vision applications, \eg visual inspection and quality management in manufacturing, cancer and tumor detection in healthcare, and vehicle re-identification and pedestrian detection in transportation.

On the other hand, the advanced vision recognition technologies could also have potential negative societal impact if they are adopted by harmful or mismanaged applications, \eg usage in surveillance systems that violate privacy. It is important to be aware when vision technologies are deployed in practical applications.

\paragraph{Limitations.}
Our \mvit is a general vision backbone and we demonstrate its effectiveness on various recognition tasks. To reduce the full hyperparameter tuning space for \mvit on different datasets and tasks, we mainly follow the existing standard recipe for each task from the community (\eg \cite{MViT,liu2021swin,deit}) with lightweight tuning (\eg learning rate, weight decay). Therefore, the choice of hyperparameters for different \mvit variants may be suboptimal. 

In addition, \mvit provides five different variants from tiny to huge models with different complexity as a general backbone. In the future, we think there are two potential interesting research directions: scaling down \mvit to even smaller models for mobile applications, and scaling up \mvit to even larger models for large-scale data scenarios.

{\small
\bibliographystyle{ieee_fullname}
\bibliography{egbib}

\begin{thebibliography}{10}\itemsep=-1pt

\bibitem{arnab2021vivit}
Anurag Arnab, Mostafa Dehghani, Georg Heigold, Chen Sun, Mario Lu{\v{c}}i{\'c},
  and Cordelia Schmid.
\newblock Vivit: A video vision transformer.
\newblock {\em arXiv preprint arXiv:2103.15691}, 2021.

\bibitem{beal2020toward}
Josh Beal, Eric Kim, Eric Tzeng, Dong~Huk Park, Andrew Zhai, and Dmitry
  Kislyuk.
\newblock Toward transformer-based object detection.
\newblock {\em arXiv preprint arXiv:2012.09958}, 2020.

\bibitem{bertasius2021space}
Gedas Bertasius, Heng Wang, and Lorenzo Torresani.
\newblock Is space-time attention all you need for video understanding?
\newblock {\em arXiv preprint arXiv:2102.05095}, 2021.

\bibitem{bodla2017soft}
Navaneeth Bodla, Bharat Singh, Rama Chellappa, and Larry~S Davis.
\newblock Soft-nms--improving object detection with one line of code.
\newblock In {\em Proc. ICCV}, 2017.

\bibitem{brock2021high}
Andrew Brock, Soham De, Samuel~L Smith, and Karen Simonyan.
\newblock High-performance large-scale image recognition without normalization.
\newblock {\em arXiv preprint arXiv:2102.06171}, 2021.

\bibitem{cai2018cascade}
Zhaowei Cai and Nuno Vasconcelos.
\newblock Cascade r-cnn: Delving into high quality object detection.
\newblock In {\em Proc. CVPR}, 2018.

\bibitem{Carreira19}
Jo{\~{a}}o Carreira, Eric Noland, Chloe Hillier, and Andrew Zisserman.
\newblock A short note on the kinetics-700 human action dataset.
\newblock {\em arXiv preprint arXiv:1907.06987}, 2019.

\bibitem{Carreira2017}
Joao Carreira and Andrew Zisserman.
\newblock Quo vadis, action recognition? a new model and the kinetics dataset.
\newblock In {\em Proc. CVPR}, 2017.

\bibitem{chen2021crossvit}
Chun-Fu Chen, Quanfu Fan, and Rameswar Panda.
\newblock Crossvit: Cross-attention multi-scale vision transformer for image
  classification.
\newblock In {\em Proc. ICCV}, 2021.

\bibitem{chen2019drop}
Yunpeng Chen, Haoqi Fang, Bing Xu, Zhicheng Yan, Yannis Kalantidis, Marcus
  Rohrbach, Shuicheng Yan, and Jiashi Feng.
\newblock Drop an octave: Reducing spatial redundancy in convolutional neural
  networks with octave convolution.
\newblock {\em arXiv preprint arXiv:1904.05049}, 2019.

\bibitem{chu2021twins}
Xiangxiang Chu, Zhi Tian, Yuqing Wang, Bo Zhang, Haibing Ren, Xiaolin Wei,
  Huaxia Xia, and Chunhua Shen.
\newblock Twins: Revisiting the design of spatial attention in vision
  transformers.
\newblock In {\em NIPS}, 2021.

\bibitem{cubuk2020randaugment}
Ekin~D Cubuk, Barret Zoph, Jonathon Shlens, and Quoc~V Le.
\newblock Randaugment: Practical automated data augmentation with a reduced
  search space.
\newblock In {\em Proc. CVPR}, 2020.

\bibitem{dai2021coatnet}
Zihang Dai, Hanxiao Liu, Quoc~V Le, and Mingxing Tan.
\newblock Coatnet: Marrying convolution and attention for all data sizes.
\newblock {\em arXiv preprint arXiv:2106.04803}, 2021.

\bibitem{deng2009imagenet}
Jia Deng, Wei Dong, Richard Socher, Li-Jia Li, Kai Li, and Li Fei-Fei.
\newblock Imagenet: A large-scale hierarchical image database.
\newblock In {\em Proc. CVPR}, pages 248--255. Ieee, 2009.

\bibitem{dollar2021fast}
Piotr Doll{\'a}r, Mannat Singh, and Ross Girshick.
\newblock Fast and accurate model scaling.
\newblock In {\em Proc. CVPR}, 2021.

\bibitem{dong2021cswin}
Xiaoyi Dong, Jianmin Bao, Dongdong Chen, Weiming Zhang, Nenghai Yu, Lu Yuan,
  Dong Chen, and Baining Guo.
\newblock Cswin transformer: A general vision transformer backbone with
  cross-shaped windows.
\newblock {\em arXiv preprint arXiv:2107.00652}, 2021.

\bibitem{dosovitskiy2020image}
Alexey Dosovitskiy, Lucas Beyer, Alexander Kolesnikov, Dirk Weissenborn,
  Xiaohua Zhai, Thomas Unterthiner, Mostafa Dehghani, Matthias Minderer, Georg
  Heigold, Sylvain Gelly, et~al.
\newblock An image is worth 16x16 words: Transformers for image recognition at
  scale.
\newblock {\em arXiv preprint arXiv:2010.11929}, 2020.

\bibitem{el2021xcit}
Alaaeldin El-Nouby, Hugo Touvron, Mathilde Caron, Piotr Bojanowski, Matthijs
  Douze, Armand Joulin, Ivan Laptev, Natalia Neverova, Gabriel Synnaeve, Jakob
  Verbeek, et~al.
\newblock Xcit: Cross-covariance image transformers.
\newblock {\em arXiv preprint arXiv:2106.09681}, 2021.

\bibitem{fan2020pyslowfast}
Haoqi Fan, Yanghao Li, Bo Xiong, Wan-Yen Lo, and Christoph Feichtenhofer.
\newblock {PySlowFast}.
\newblock \url{https://github.com/facebookresearch/slowfast}, 2020.

\bibitem{fan2021pytorchvideo}
Haoqi Fan, Tullie Murrell, Heng Wang, Kalyan~Vasudev Alwala, Yanghao Li, Yilei
  Li, Bo Xiong, Nikhila Ravi, Meng Li, Haichuan Yang, Jitendra Malik, Ross
  Girshick, Matt Feiszli, Aaron Adcock, Wan-Yen Lo, and Christoph
  Feichtenhofer.
\newblock {PyTorchVideo}: A deep learning library for video understanding.
\newblock In {\em Proceedings of the 29th ACM International Conference on
  Multimedia}, 2021.
\newblock \url{https://pytorchvideo.org/}.

\bibitem{MViT}
Haoqi Fan, Bo Xiong, Karttikeya Mangalam, Yanghao Li, Zhicheng Yan, Jitendra
  Malik, and Christoph Feichtenhofer.
\newblock Multiscale vision transformers.
\newblock In {\em Proc. ICCV}, 2021.

\bibitem{feichtenhofer2020x3d}
Christoph Feichtenhofer.
\newblock {X3D}: Expanding architectures for efficient video recognition.
\newblock In {\em Proc. CVPR}, pages 203--213, 2020.

\bibitem{Feichtenhofer2019}
Christoph Feichtenhofer, Haoqi Fan, Jitendra Malik, and Kaiming He.
\newblock {SlowFast} networks for video recognition.
\newblock In {\em Proc. ICCV}, 2019.

\bibitem{Feichtenhofer2016a}
Christoph Feichtenhofer, Axel Pinz, and Richard Wildes.
\newblock Spatiotemporal residual networks for video action recognition.
\newblock In {\em NIPS}, 2016.

\bibitem{Feichtenhofer2016}
Christoph Feichtenhofer, Axel Pinz, and Andrew Zisserman.
\newblock Convolutional two-stream network fusion for video action recognition.
\newblock In {\em Proc. CVPR}, 2016.

\bibitem{ghiasi2021simple}
Golnaz Ghiasi, Yin Cui, Aravind Srinivas, Rui Qian, Tsung-Yi Lin, Ekin~D Cubuk,
  Quoc~V Le, and Barret Zoph.
\newblock Simple copy-paste is a strong data augmentation method for instance
  segmentation.
\newblock In {\em Proc. CVPR}, 2021.

\bibitem{ghiasi2019fpn}
Golnaz Ghiasi, Tsung-Yi Lin, and Quoc~V Le.
\newblock Nas-fpn: Learning scalable feature pyramid architecture for object
  detection.
\newblock In {\em Proc. CVPR}, 2019.

\bibitem{girdhar2019video}
Rohit Girdhar, Joao Carreira, Carl Doersch, and Andrew Zisserman.
\newblock Video action transformer network.
\newblock In {\em Proc. CVPR}, 2019.

\bibitem{Girshick2015}
Ross Girshick.
\newblock {Fast R-CNN}.
\newblock In {\em Proc. ICCV}, 2015.

\bibitem{Goyal2017}
Priya Goyal, Piotr Doll{\'a}r, Ross Girshick, Pieter Noordhuis, Lukasz
  Wesolowski, Aapo Kyrola, Andrew Tulloch, Yangqing Jia, and Kaiming He.
\newblock {Accurate, large minibatch SGD: training ImageNet in 1 hour}.
\newblock {\em arXiv:1706.02677}, 2017.

\bibitem{ssv2}
Raghav Goyal, Samira~Ebrahimi Kahou, Vincent Michalski, Joanna Materzynska,
  Susanne Westphal, Heuna Kim, Valentin Haenel, Ingo Fruend, Peter Yianilos,
  Moritz Mueller-Freitag, et~al.
\newblock The {``Something Something"} video database for learning and
  evaluating visual common sense.
\newblock In {\em ICCV}, 2017.

\bibitem{Gu2018}
Chunhui Gu, Chen Sun, David~A. Ross, Carl Vondrick, Caroline Pantofaru, Yeqing
  Li, Sudheendra Vijayanarasimhan, George Toderici, Susanna Ricco, Rahul
  Sukthankar, Cordelia Schmid, and Jitendra Malik.
\newblock {AVA}: A video dataset of spatio-temporally localized atomic visual
  actions.
\newblock In {\em Proc. CVPR}, 2018.

\bibitem{han2021tnt}
Kai Han, An Xiao, Enhua Wu, Jianyuan Guo, Chunjing Xu, and Yunhe Wang.
\newblock Transformer in transformer.
\newblock In {\em NIPS}, 2021.

\bibitem{resnest}
Zhang Hang, Chongruo Wu, Zhongyue Zhang, Yi Zhu, Zhi Zhang, Haibin Lin, and Yue
  Sun.
\newblock Resnest: Split-attention networks.
\newblock 2020.

\bibitem{hanin2018start}
Boris Hanin and David Rolnick.
\newblock How to start training: The effect of initialization and architecture.
\newblock {\em arXiv preprint arXiv:1803.01719}, 2018.

\bibitem{He2017}
Kaiming He, Georgia Gkioxari, Piotr Doll{\'a}r, and Ross Girshick.
\newblock Mask {R-CNN}.
\newblock In {\em Proc. ICCV}, 2017.

\bibitem{He2015}
Kaiming He, Xiangyu Zhang, Shaoqing Ren, and Jian Sun.
\newblock Delving deep into rectifiers: Surpassing human-level performance on
  imagenet classification.
\newblock In {\em Proc. CVPR}, 2015.

\bibitem{He2016}
Kaiming He, Xiangyu Zhang, Shaoqing Ren, and Jian Sun.
\newblock Deep residual learning for image recognition.
\newblock In {\em Proc. CVPR}, 2016.

\bibitem{He2016a}
Kaiming He, Xiangyu Zhang, Shaoqing Ren, and Jian Sun.
\newblock Identity mappings in deep residual networks.
\newblock In {\em Proc. ECCV}, 2016.

\bibitem{hoffer2020augment}
Elad Hoffer, Tal Ben-Nun, Itay Hubara, Niv Giladi, Torsten Hoefler, and Daniel
  Soudry.
\newblock Augment your batch: Improving generalization through instance
  repetition.
\newblock In {\em Proc. CVPR}, pages 8129--8138, 2020.

\bibitem{huang2016deep}
Gao Huang, Yu Sun, Zhuang Liu, Daniel Sedra, and Kilian~Q Weinberger.
\newblock Deep networks with stochastic depth.
\newblock In {\em Proc. ECCV}, 2016.

\bibitem{jiang2019stm}
Boyuan Jiang, MengMeng Wang, Weihao Gan, Wei Wu, and Junjie Yan.
\newblock Stm: Spatiotemporal and motion encoding for action recognition.
\newblock In {\em Proc. CVPR}, pages 2000--2009, 2019.

\bibitem{jiang2021token}
Zihang Jiang, Qibin Hou, Li Yuan, Daquan Zhou, Xiaojie Jin, Anran Wang, and
  Jiashi Feng.
\newblock Token labeling: Training a 85.5\% top-1 accuracy vision transformer
  with 56m parameters on imagenet.
\newblock {\em arXiv preprint arXiv:2104.10858}, 2021.

\bibitem{Kay2017}
Will Kay, Joao Carreira, Karen Simonyan, Brian Zhang, Chloe Hillier, Sudheendra
  Vijayanarasimhan, Fabio Viola, Tim Green, Trevor Back, Paul Natsev, et~al.
\newblock The kinetics human action video dataset.
\newblock {\em arXiv:1705.06950}, 2017.

\bibitem{kondratyuk2021movinets}
Dan Kondratyuk, Liangzhe Yuan, Yandong Li, Li Zhang, Mingxing Tan, Matthew
  Brown, and Boqing Gong.
\newblock {MoViNets}: Mobile video networks for efficient video recognition.
\newblock In {\em Proc. CVPR}, 2021.

\bibitem{Krizhevsky2012}
Alex Krizhevsky, Ilya Sutskever, and Geoffrey~E Hinton.
\newblock {ImageNet} classification with deep convolutional neural networks.
\newblock In {\em NIPS}, 2012.

\bibitem{lecun1989handwritten}
Yann LeCun, Bernhard Boser, John Denker, Donnie Henderson, Richard Howard,
  Wayne Hubbard, and Lawrence Jackel.
\newblock Handwritten digit recognition with a back-propagation network.
\newblock In {\em NIPS}, 1989.

\bibitem{lecun1989backpropagation}
Yann LeCun, Bernhard Boser, John~S Denker, Donnie Henderson, Richard~E Howard,
  Wayne Hubbard, and Lawrence~D Jackel.
\newblock Backpropagation applied to handwritten zip code recognition.
\newblock {\em Neural computation}, 1(4):541--551, 1989.

\bibitem{li2020tea}
Yan Li, Bin Ji, Xintian Shi, Jianguo Zhang, Bin Kang, and Limin Wang.
\newblock Tea: Temporal excitation and aggregation for action recognition.
\newblock In {\em Proc. CVPR}, pages 909--918, 2020.

\bibitem{li2021benchmarking}
Yanghao Li, Saining Xie, Xinlei Chen, Piotr Dollar, Kaiming He, and Ross
  Girshick.
\newblock Benchmarking detection transfer learning with vision transformers.
\newblock {\em arXiv preprint arXiv:2111.11429}, 2021.

\bibitem{Li2018}
Zhenyang Li, Kirill Gavrilyuk, Efstratios Gavves, Mihir Jain, and Cees~GM
  Snoek.
\newblock {VideoLSTM} convolves, attends and flows for action recognition.
\newblock {\em Computer Vision and Image Understanding}, 166:41--50, 2018.

\bibitem{lin2018temporal}
Ji Lin, Chuang Gan, and Song Han.
\newblock Temporal shift module for efficient video understanding.
\newblock In {\em Proc. ICCV}, 2019.

\bibitem{Lin2017}
Tsung-Yi Lin, Piotr Doll{\'a}r, Ross Girshick, Kaiming He, Bharath Hariharan,
  and Serge Belongie.
\newblock Feature pyramid networks for object detection.
\newblock In {\em Proc. CVPR}, 2017.

\bibitem{Lin2014}
Tsung-Yi Lin, Michael Maire, Serge Belongie, James Hays, Pietro Perona, Deva
  Ramanan, Piotr Doll{\'a}r, and C~Lawrence Zitnick.
\newblock Microsoft {COCO}: Common objects in context.
\newblock In {\em Proc. ECCV}, 2014.

\bibitem{liu2021swin}
Ze Liu, Yutong Lin, Yue Cao, Han Hu, Yixuan Wei, Zheng Zhang, Stephen Lin, and
  Baining Guo.
\newblock Swin transformer: Hierarchical vision transformer using shifted
  windows.
\newblock {\em arXiv preprint arXiv:2103.14030}, 2021.

\bibitem{liu2021video}
Ze Liu, Jia Ning, Yue Cao, Yixuan Wei, Zheng Zhang, Stephen Lin, and Han Hu.
\newblock Video swin transformer.
\newblock {\em arXiv preprint arXiv:2106.13230}, 2021.

\bibitem{Loshchilov2016}
Ilya Loshchilov and Frank Hutter.
\newblock {SGDR}: Stochastic gradient descent with warm restarts.
\newblock {\em arXiv:1608.03983}, 2016.

\bibitem{loshchilov2018fixing}
Ilya Loshchilov and Frank Hutter.
\newblock Fixing weight decay regularization in adam.
\newblock 2018.

\bibitem{neimark2021video}
Daniel Neimark, Omri Bar, Maya Zohar, and Dotan Asselmann.
\newblock Video transformer network.
\newblock {\em arXiv preprint arXiv:2102.00719}, 2021.

\bibitem{pan2021actor}
Junting Pan, Siyu Chen, Mike~Zheng Shou, Yu Liu, Jing Shao, and Hongsheng Li.
\newblock Actor-context-actor relation network for spatio-temporal action
  localization.
\newblock In {\em Proc. CVPR}, 2021.

\bibitem{Qiu2017}
Zhaofan Qiu, Ting Yao, and Tao Mei.
\newblock Learning spatio-temporal representation with pseudo-3d residual
  networks.
\newblock In {\em Proc. ICCV}, 2017.

\bibitem{ilija_2020}
Ilija Radosavovic, Raj Prateek~Kosaraju, Ross Girshick, Kaiming He, and Piotr
  Dollár.
\newblock Designing network design spaces.
\newblock In {\em Proc. CVPR}, June 2020.

\bibitem{Redmon2016}
Joseph Redmon, Santosh Divvala, Ross Girshick, and Ali Farhadi.
\newblock You only look once: Unified, real-time object detection.
\newblock In {\em Proc. CVPR}, 2016.

\bibitem{Ren2015}
Shaoqing Ren, Kaiming He, Ross Girshick, and Jian Sun.
\newblock {Faster R-CNN}: Towards real-time object detection with region
  proposal networks.
\newblock In {\em NIPS}, 2015.

\bibitem{shaw2018self}
Peter Shaw, Jakob Uszkoreit, and Ashish Vaswani.
\newblock Self-attention with relative position representations.
\newblock {\em arXiv preprint arXiv:1803.02155}, 2018.

\bibitem{Simonyan2014}
Karen Simonyan and Andrew Zisserman.
\newblock Two-stream convolutional networks for action recognition in videos.
\newblock In {\em NIPS}, 2014.

\bibitem{Simonyan2015}
Karen Simonyan and Andrew Zisserman.
\newblock Very deep convolutional networks for large-scale image recognition.
\newblock In {\em Proc. ICLR}, 2015.

\bibitem{strudel2021segmenter}
Robin Strudel, Ricardo Garcia, Ivan Laptev, and Cordelia Schmid.
\newblock Segmenter: Transformer for semantic segmentation.
\newblock {\em arXiv preprint arXiv:2105.05633}, 2021.

\bibitem{Szegedy2015}
Christian Szegedy, Wei Liu, Yangqing Jia, Pierre Sermanet, Scott Reed, Dragomir
  Anguelov, Dumitru Erhan, Vincent Vanhoucke, and Andrew Rabinovich.
\newblock Going deeper with convolutions.
\newblock In {\em Proc. CVPR}, 2015.

\bibitem{Szegedy2015a}
Christian Szegedy, Vincent Vanhoucke, Sergey Ioffe, Jonathon Shlens, and
  Zbigniew Wojna.
\newblock Rethinking the inception architecture for computer vision.
\newblock {\em arXiv:1512.00567}, 2015.

\bibitem{tan2019efficientnet}
Mingxing Tan and Quoc~V Le.
\newblock Efficientnet: Rethinking model scaling for convolutional neural
  networks.
\newblock {\em arXiv preprint arXiv:1905.11946}, 2019.

\bibitem{touvron2020training}
Hugo Touvron, Matthieu Cord, Matthijs Douze, Francisco Massa, Alexandre
  Sablayrolles, and Herv{\'e} J{\'e}gou.
\newblock Training data-efficient image transformers \& distillation through
  attention.
\newblock {\em arXiv preprint arXiv:2012.12877}, 2020.

\bibitem{deit}
Hugo Touvron, Matthieu Cord, Matthijs Douze, Francisco Massa, Alexandre
  Sablayrolles, and Hervé Jégou.
\newblock {DeiT}: Data-efficient image transformers.
\newblock {\em arXiv preprint arXiv:2012.12877}, 2020.

\bibitem{touvron2021going}
Hugo Touvron, Matthieu Cord, Alexandre Sablayrolles, Gabriel Synnaeve, and
  Herv{\'e} J{\'e}gou.
\newblock Going deeper with image transformers.
\newblock {\em arXiv preprint arXiv:2103.17239}, 2021.

\bibitem{Tran2019}
Du Tran, Heng Wang, Lorenzo Torresani, and Matt Feiszli.
\newblock Video classification with channel-separated convolutional networks.
\newblock In {\em Proc. ICCV}, 2019.

\bibitem{vaswani2017attention}
Ashish Vaswani, Noam Shazeer, Niki Parmar, Jakob Uszkoreit, Llion Jones,
  Aidan~N Gomez, Lukasz Kaiser, and Illia Polosukhin.
\newblock Attention is all you need.
\newblock {\em arXiv preprint arXiv:1706.03762}, 2017.

\bibitem{wang2021pvtv2}
Wenhai Wang, Enze Xie, Xiang Li, Deng-Ping Fan, Kaitao Song, Ding Liang, Tong
  Lu, Ping Luo, and Ling Shao.
\newblock Pvtv2: Improved baselines with pyramid vision transformer.
\newblock {\em arXiv preprint arXiv:2106.13797}, 2021.

\bibitem{wang2021pyramid}
Wenhai Wang, Enze Xie, Xiang Li, Deng-Ping Fan, Kaitao Song, Ding Liang, Tong
  Lu, Ping Luo, and Ling Shao.
\newblock Pyramid vision transformer: A versatile backbone for dense prediction
  without convolutions.
\newblock In {\em IEEE ICCV}, 2021.

\bibitem{Wu2019}
Chao-Yuan Wu, Christoph Feichtenhofer, Haoqi Fan, Kaiming He, Philipp
  Kr\"{a}henb\"{u}hl, and Ross Girshick.
\newblock Long-term feature banks for detailed video understanding.
\newblock In {\em Proc. CVPR}, 2019.

\bibitem{wu2021towards}
Chao-Yuan Wu and Philipp Krahenbuhl.
\newblock Towards long-form video understanding.
\newblock In {\em Proc. CVPR}, 2021.

\bibitem{wu2021cvt}
Haiping Wu, Bin Xiao, Noel Codella, Mengchen Liu, Xiyang Dai, Lu Yuan, and Lei
  Zhang.
\newblock Cvt: Introducing convolutions to vision transformers.
\newblock {\em arXiv preprint arXiv:2103.15808}, 2021.

\bibitem{wu2019detectron2}
Yuxin Wu, Alexander Kirillov, Francisco Massa, Wan-Yen Lo, and Ross Girshick.
\newblock Detectron2.
\newblock \url{https://github.com/facebookresearch/detectron2}, 2019.

\bibitem{Xie2017}
Saining Xie, Ross Girshick, Piotr Doll{\'a}r, Zhuowen Tu, and Kaiming He.
\newblock Aggregated residual transformations for deep neural networks.
\newblock In {\em Proc. CVPR}, 2017.

\bibitem{Xie2018}
Saining Xie, Chen Sun, Jonathan Huang, Zhuowen Tu, and Kevin Murphy.
\newblock Rethinking spatiotemporal feature learning for video understanding.
\newblock {\em arXiv:1712.04851}, 2017.

\bibitem{yuan2021tokens}
Li Yuan, Yunpeng Chen, Tao Wang, Weihao Yu, Yujun Shi, Zi-Hang Jiang,
  Francis~E.H. Tay, Jiashi Feng, and Shuicheng Yan.
\newblock Tokens-to-token vit: Training vision transformers from scratch on
  imagenet.
\newblock In {\em Proc. ICCV}, 2021.

\bibitem{yuan2021volo}
Li Yuan, Qibin Hou, Zihang Jiang, Jiashi Feng, and Shuicheng Yan.
\newblock Volo: Vision outlooker for visual recognition, 2021.

\bibitem{yun2019cutmix}
Sangdoo Yun, Dongyoon Han, Seong~Joon Oh, Sanghyuk Chun, Junsuk Choe, and
  Youngjoon Yoo.
\newblock Cutmix: Regularization strategy to train strong classifiers with
  localizable features.
\newblock In {\em Proc. ICCV}, 2019.

\bibitem{zhang2018mixup}
Hongyi Zhang, Moustapha Cisse, Yann~N Dauphin, and David Lopez-Paz.
\newblock Mixup: Beyond empirical risk minimization.
\newblock In {\em Proc. ICLR}, 2018.

\bibitem{zhang2021multi}
Pengchuan Zhang, Xiyang Dai, Jianwei Yang, Bin Xiao, Lu Yuan, Lei Zhang, and
  Jianfeng Gao.
\newblock Multi-scale vision longformer: A new vision transformer for
  high-resolution image encoding.
\newblock In {\em Proc. ICCV}, 2021.

\bibitem{zheng2021rethinking}
Sixiao Zheng, Jiachen Lu, Hengshuang Zhao, Xiatian Zhu, Zekun Luo, Yabiao Wang,
  Yanwei Fu, Jianfeng Feng, Tao Xiang, Philip~HS Torr, et~al.
\newblock Rethinking semantic segmentation from a sequence-to-sequence
  perspective with transformers.
\newblock In {\em Proc. CVPR}, 2021.

\bibitem{zhong2020random}
Zhun Zhong, Liang Zheng, Guoliang Kang, Shaozi Li, and Yi Yang.
\newblock Random erasing data augmentation.
\newblock In {\em Proceedings of the AAAI Conference on Artificial
  Intelligence}, volume~34, pages 13001--13008, 2020.

\bibitem{Zhou2017}
Bolei Zhou, Alex Andonian, Aude Oliva, and Antonio Torralba.
\newblock Temporal relational reasoning in videos.
\newblock In {\em ECCV}, 2018.

\bibitem{zhou2019objects}
Xingyi Zhou, Dequan Wang, and Philipp Kr{\"a}henb{\"u}hl.
\newblock Objects as points.
\newblock {\em arXiv preprint arXiv:1904.07850}, 2019.

\end{thebibliography}
}

\end{document}